\documentclass[12pt]{article}

\usepackage{scicite}

\usepackage{times}
\usepackage{amsmath,amsfonts}
\usepackage{algorithmic}
\usepackage{algorithm}
\usepackage[algo2e]{algorithm2e} 
\usepackage{array}
\usepackage[caption=false,font=normalsize,labelfont=sf,textfont=sf]{subfig}
\usepackage{textcomp}
\usepackage{stfloats}
\usepackage{url}
\usepackage{verbatim}
\usepackage{graphicx}

\newcommand{\RNum}[1]{\uppercase\expandafter{\romannumeral #1\relax}}
\newcommand*{\B}[1]{\ifmmode\bm{#1}\else\textbf{#1}\fi}

\DeclareMathOperator{\FFN}{FFN}
\DeclareMathOperator{\ConvLSTM}{ConvLSTM}
\DeclareMathOperator{\LSTM}{LSTM}

\DeclareMathOperator{\ATT}{ATT}
\DeclareMathOperator{\TRAN}{TRAN}
\DeclareMathOperator{\deconv}{Deconv}
\newcommand{\VWMoneSuffix}{\text{M1}}
\newcommand{\VWMtwoSuffix}{\text{M2}}
\usepackage{algorithm2e}  
\usepackage{svg}    

\usepackage{color}
\usepackage{booktabs}
\usepackage{accents}
\usepackage{multirow}
\usepackage{rotating}
\usepackage{bbm}
\usepackage{latexsym}
\usepackage{bm}
\usepackage{xfrac}
\usepackage{hyperref}
\hypersetup{}

\newenvironment{sciabstract}{%
\begin{quote} \bf}
{\end{quote}}

\title{Development of Compositionality and \\ Generalization through Interactive Learning\\ of Language and Action of Robots} 
\author
{Prasanna Vijayaraghavan,$^{1}$ Jeffrey Frederic Quei{\ss}er,$^{1}$\\
Sergio Verduzco Flores,$^{1}$ Jun Tani$^{1\ast}$ \\
\\
\normalsize{$^{1}$Okinawa Institute of Science and Technology,}\\
\normalsize{Okinawa, Japan}\\
\\
\normalsize{$^\ast$To whom correspondence should be addressed; E-mail:  jun.tani@oist.jp.}
}

\date{}

\begin{document} 


\baselineskip24pt

%
\maketitle 


\begin{abstract}
    \textbf{One Sentence Summary}: Generalization in learning verb-noun compositions improves significantly with increased training task variations. 
\end{abstract}

\begin{sciabstract}
Humans excel at applying learned behavior to unlearned situations. A crucial component of this generalization behavior is our ability to compose/decompose a whole into reusable parts, an attribute known as compositionality. One of the fundamental questions in robotics concerns this characteristic. "How can linguistic compositionality be developed concomitantly with sensorimotor skills through associative learning, particularly when individuals only learn partial linguistic compositions and their corresponding sensorimotor patterns?" To address this question, we propose a brain-inspired neural network model that integrates vision, proprioception, and language into a framework of predictive coding and active inference, based on the free-energy principle. The effectiveness and capabilities of this model were assessed through various simulation experiments conducted with a robot arm. Our results show that generalization in learning to unlearned verb-noun compositions, is significantly enhanced when training variations of task composition are increased. We attribute this to self-organized compositional structures in linguistic latent state space being influenced significantly by sensorimotor learning. Ablation studies show that visual attention and working memory are essential to accurately generate visuo-motor sequences to achieve linguistically represented goals. These insights advance our understanding of mechanisms underlying development of compositionality through interactions of linguistic and sensorimotor experience. 
\end{sciabstract}

\section*{Introduction}
The problem of generalizing learned behavior to unlearned situations is easy for humans, but incredibly challenging for cognitive robots. Compositionality \cite{Chomsky1957,Evans1982,Frege1991-FRECPO,Janssen2001} is a major linguistic competency that is essential for generalization of cognitive behavior. Lake and colleagues \cite{lake_ullman_tenenbaum_gershman_2017} consider compositionality as one of the three fundamental competencies necessary to build machines that can learn to think like humans. Although interpretations of compositionality vary, Hupkes et.al. \cite{hupkes2020} define it by identifying its essential components. Among these, \textit{systematicity}, the ability to recombine known parts and rules for use in a novel context, is a central component of compositionality, on which we focus in the current study. Recent deep learning models seem to trivialize this problem, but in reality, they offer little insight into how language develops in humans. Although it can be argued that using large language models (LLMs) for end-to-end learning shows that they can understand the meanings of words by learning from a large corpus collected in the real world \cite{lynch2022interactive, nolfi2023unexpected, abdou2021language, yousefi2024decoding, pavlick2023}, they cannot access any sensorimotor patterns associated with words and sentences. Our objective is to understand how the aforementioned systematicity aspect of compositionality in language and behavior can co-develop through their interactions, by building an integrative neural network model for conducting robotic simulation experiments.

We use a developmental robotic approach in conjunction with the free-energy principle (FEP) \cite{FristonFEP} to address this problem. Modeling embodied language with developmental robotics is consistent with the constructivist view, or usage-based theory of language acquisition \cite{Cameron-Faulkner2003,tomasello2009usage}. Embodiment is considered a necessary precondition for developing higher thoughts \cite{Smith2005}. According to Piaget \cite{piaget_language_1955} infants develop body-rationality representation through sensorimotor interactions with the environment, accompanied by goal-directed actions. Neuroscience researchers \cite{BUCCINO2005355, DREYER201852,Pulvermuller2010} have found that modulation of motor system activity occurs while listening to sentences expressing actions, suggesting that humans infer actions from language and vice-versa. Developmental robotics \cite{Oudeyer2019} also addresses the {\it symbol grounding} \cite{harnad1990} problem, which seeks to understand how symbols commonly used in linguistic expressions are associated with meaning in the real world. Cognitive competencies such as visual attention and visual working memory (VWM) are crucial in development of embodied language \cite{tomasello2006social,tomasello1992first,tomasello2009usage,Smith2018,Cangelosi2022}.
Using developmental robotics, several studies have investigated the association of language and visuo-proprioceptive behavior with hierarchical multi-modal recurrent neural networks\cite{Sugita2005,Cangelosi2010,Cangelosi2018,Heinrich2018,akakzia2021decstr, yamada2018paired}. 

In parallel with the developmental robotic approach, recent advances in cognitive neuroscience have underscored the significance of theoretical frameworks such as the free-energy principle (FEP) in modeling cognitive brain mechanisms. According to the FEP, perception and action are modeled in the framework of predictive coding (PC)\cite{rao1999,friston2009PC} and active inference (AIF)\cite{friston2009,friston2010,brown_friston_bestmann_2011}, respectively. PC is a theory of perception that provides a unifying framework for neuronal mechanisms of top-down prediction and perceptual learning of sensory information. AIF is a process for inferring actions that minimize the error between preferred sensation and predicted actional outcomes. Some neural network models \cite{Matsumoto2020,Matsumoto2022,jeff2021,friston2010,brown_friston_bestmann_2011} have been successfully incorporated into goal-directed planning schemes based on active inference to show that artificial agents can generate adequate goal-directed behaviors based on learning in the habituated range of the world. Some of these models \cite{friston2010, brown_friston_bestmann_2011} generated action plans by optimizing the policy, and others \cite{jeff2021, Matsumoto2020, Matsumoto2022} optimize low-dimensional latent variables by minimizing the future expected free energy. Teleology, a philosophical concept that explains phenomena based on their ultimate goals or purposes rather than merely their causes or origins, aligns closely with principles of goal-directed behavior observed in these models. In the context of human behavior, teleology offers a framework for interpreting actions as inherently goal-directed and purpose-driven \cite{sehon2007goal,csibra2003one}, providing a philosophical underpinning that complements mechanistic insights offered by FEP-based approaches.

Inspired by these ideas we propose a neural network model (Figure 1A), to study co-development of linguistic compositionality paralleling sensorimotor experience. It consists of RNN-based generative networks that handle prediction of vision, proprioception, and language. These modalities are integrated by the associative network. Our model utilizes an executive layer mechanized by predictive coding inspired variational RNN (PV-RNN) \cite{ahmadi2019pvrnn} to integrate language with visuo-proprioceptive sequences. PV-RNN, a neural network consistent with the FEP, contains probabilistic latent variables that allow it to learn probabilistic structure hidden in the data. Some studies \cite{BERNARDI2020954, NEURIPS2022_d0241a0f} have found that abstract representations such as vector representations reconcile compositional generalization with distributed neural codes. 

We introduce a parametric bias ($\mathbf{PB}$) \cite{tani2003,tani2004} vector as the language latent variable. $\mathbf{PB}$ is a low dimensional latent state vector used in recurrent neural network models. In training of multiple temporal patterns, $\mathbf{PB}$ vector space is self-organized such that each temporal pattern is encoded by a specific  point in the $\mathbf{PB}$ vector space. Sugita and Tani \cite{Sugita2005} showed that word sequences and corresponding behavioral temporal patterns can be bound using the PB facilitating the development of linguistic compositionality. In the current model, the associative PV-RNN is constrained by the $\mathbf{PB}$ vector, which influences learning of associations between vision, proprioception, and linguistics.

This model learns to generate visuo-proprioceptive sequences with appropriate linguistic predictions by minimizing evidence free energy. The trained model generates appropriate visuo-proprioceptive sequences to achieve linguistically represented goals via goal-directed planning, by means of active inference. Figure 1(B) and (C) show the visuo-motor sequences predicted by the model compared with observed ground truth. We use a teleology-inspired approach to goal-directed planning proposed by Matsumoto et.al \cite{Matsumoto2022}. The underlying concept is that goal expectation is generated at every time step instead of expecting the goal at a distal step. 

Through simulation experiments we reach the following conclusions: First, generalization in learning improves significantly as the number of variations in task compositions increase. 
Second, the compositional structure that emerges in the linguistic latent state representation is significantly influenced by sensorimotor learning. Specifically, we observed that the linguistic latent representation of actional concepts develops by preserving similarity among corresponding sensorimotor patterns. Last, by performing ablation studies we found that the model's ability to accurately generate visuo-proprioceptive sequences is significantly impacted by the presence of visual attention and working memory modules.

\begin{figure}[tbph]
\centering
\includegraphics[width=0.9\textwidth]{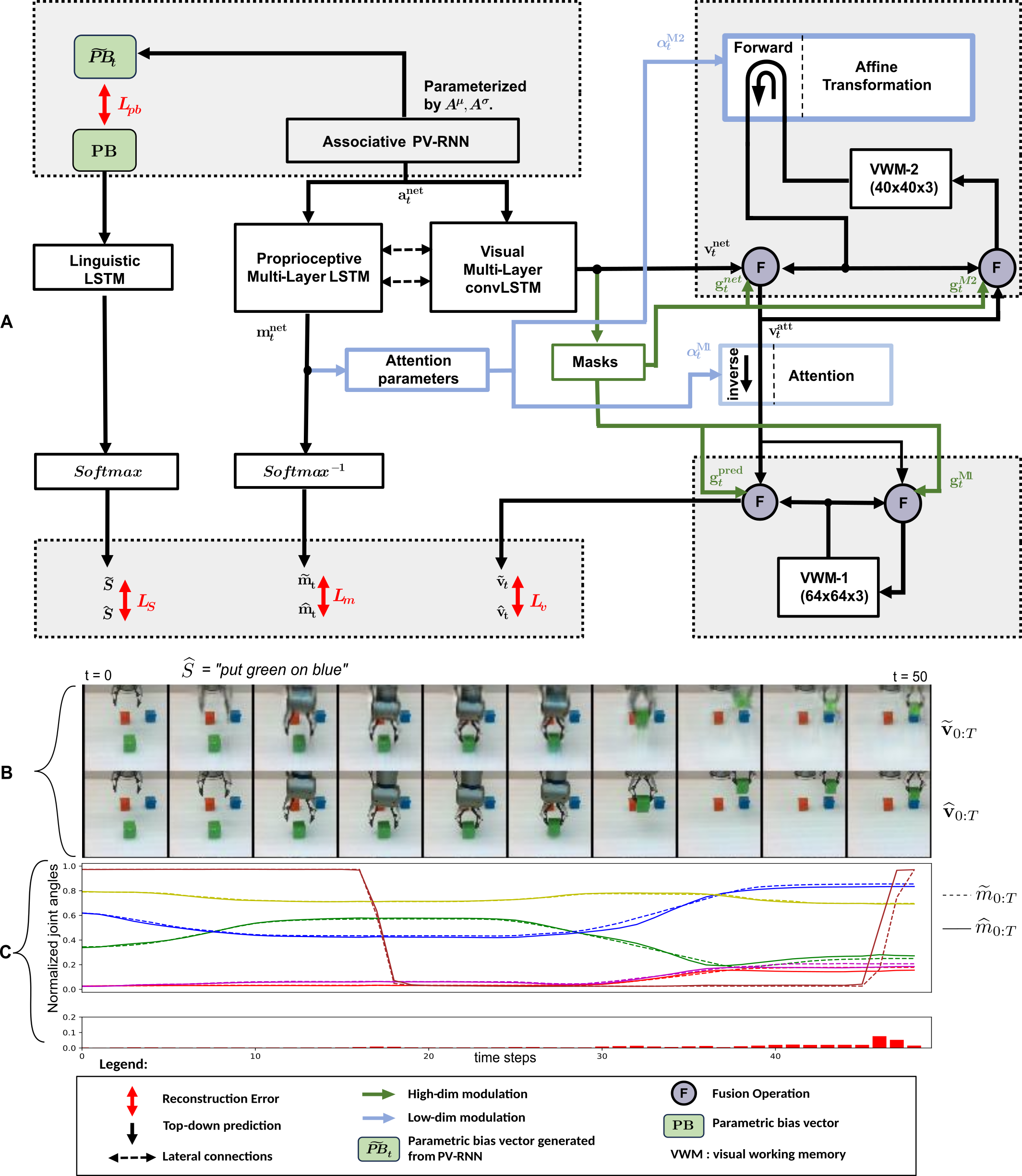}
\caption{(\textbf{A}) Model Architecture: Each modality generates visual, proprioceptive or linguistic predictions. Visual (conv-LSTM) and Proprioceptive (LSTM) modalities are integrated by the Associative PV-RNN and Linguistic LSTM is bound to Associative PV-RNN via Parametric Bias (PB). Visual predictions are enhanced by two visual working memory (VWM-1 amd VWM-2) and attention mechanism, for which the parameters are generated by the Proprioceptive LSTM. For a given linguistic goal \textit{"put green on blue"}, (\textbf{B}) top: predicted visual sequence, bottom: observed ground truth; (\textbf{C}) top: joint angle trajectory predicted by the model compared with the ground truth, bottom: motor prediction error.} 
\label{overall_architecture}
\end{figure}

\section*{Results}

In the current study we introduce vision based object manipulation tasks with a robotic arm (Figure S1). The tasks include grasping, moving (in four different directions; left, right, front, and back) and stacking. These tasks were performed on 5-cm cubic blocks of five colors  (\textit{red, green, blue, purple} and \textit{yellow}). Tasks are linguistically represented by sentences like \textit{"grasp X ."}, \textit{"move X left ."},\textit{ "move X right ."}, \textit{"move X front ."}, \textit{"move X back ."} and \textit{"put X on Y ."}; \textit{"X"} indicates the color of the object being manipulated (any of the five colors) and \textit{"Y"} is the color of the object at the base (we use \textit{green, blue} or \textit{yellow}) in the stacking task. Examples of visuo-motor sequences of different types of tasks are shown in Figure S2. In total, there are forty possible combinations (5 nouns, 8 verbs). The model was trained with data collected from the physical robot. However, all evaluations were performed based on the ability of the model to generate mentally simulated trajectories of visuo-proprioceptive sequences.

We perform two experimental evaluations with the above setup. First, we evaluate model performance for generalization to unlearned object positions and unlearned linguistic compositions. We further emphasize this by comparing model performance among different degrees of sparsity in training data. We also evaluate the model's ability to understand visuo-proprioceptive behavioral sequences by inferring the appropriate linguistic description. Second, we perform an ablation experiment in order to study the impact of visual attention and working memory on the model's generalization capability. 

\subsection*{Experiment I}
In this experiment we evaluated the ability of the model to generalize to unlearned object positions and unlearned language compositions. The dataset was divided into four groups, each with a different number of combinations. Group A contained 40 combinations (5 nouns and 8 verbs). Group B comprised 30 combinations (5 nouns and 6 verbs). Group C included 15 combinations (5 nouns and 3 verbs) and Group D contained 9 combinations (3 nouns and 3 verbs). In order to evaluate model performance in generalization in learning, we further divided each group with different ratios of training. Details of different training ratios in each group are described in Table 1. Note that in Group D, since the total number of combinations is nine, we used the ratios 77\%, 66\% and 33\% instead of 80\%, 60\% and 40\%. Details of individual compositions in the four groups are illustrated in Figures S3, S4 and S5.

\begin{table}[htbp]
    \centering
    \caption{Training ratios}
    \begin{tabular}{c ccc}
        \hline
            \bf{Groups}   &\bf{1}  &\bf{2} &\bf{3} \\
        \hline
            \bf{Group A (5x8)}   & 32/40 ($80\%)$  & 24/40 ($60\%$)  & 16/40( $40\%)$      \\ 
            \bf{Group B (5x6)}   & 24/30 ($80\%$)  & 18/30 ($60\%$)  & 12/30 ($40\%)$      \\ 
            \bf{Group C (5x3)}   & 12/15 ($80\%$)  & 9/15 ($60\%$)  & 6/15 ($40\%$)      \\ 
            \bf{Group D (3x3)}   & 7/9 ($77\%$)  & 6/9 ($66\%$)  & 3/9 $(33\%$)      \\ 
        \hline
    \end{tabular}
    \label{training ratio table}
\end{table}

\subsubsection*{Inference of Visuo-Proprioceptive Sequences to Achieve Linguistically Specified Goals}

The model was trained with visuo-proprioceptive sequences that start at random initial configurations of objects in the work space. As previously mentioned, the trained model performed goal-directed planning using active inference for novel object positions and unlearned compositions of linguistically represented goals (U-C), as well as for novel object positions and learned linguistically represented goals (U-P). The error between the visuo-proprioceptive plan generated by the network and the ground truth of visuo-proprioceptive trajectory was measured to evaluate the model's generalization performance. 

\begin{figure*}[tbph]
\centering
\includegraphics[width=1\textwidth]{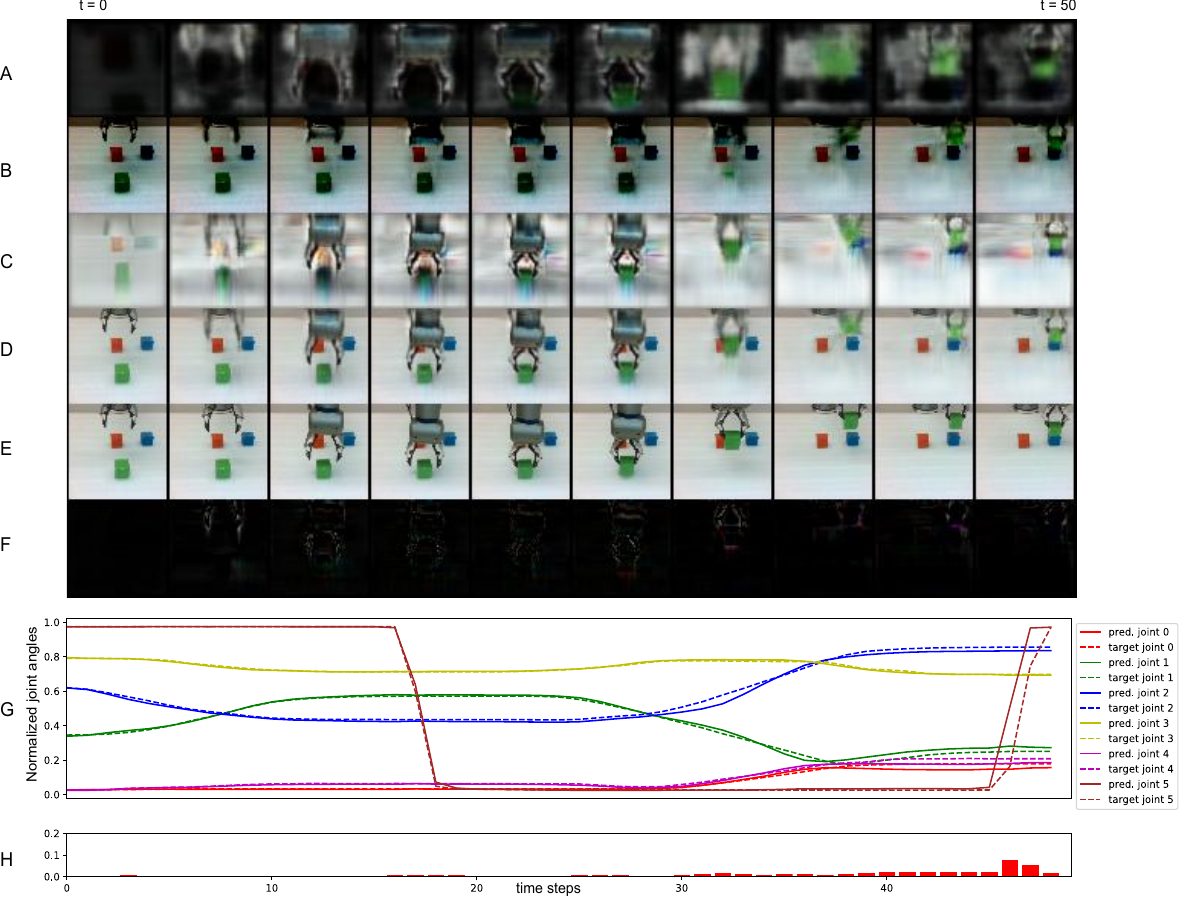}
\caption{Goal-directed planning using active inference: the model generated the above visuo-proprioceptive sequence for the linguistically specified goal \textit{"put green on blue ."}. (\textbf{A}) masked representaiton of VWM-2; (\textbf{B}) VWM-1; (\textbf{C}) model prediction of attended visual stream; (\textbf{D}) final simulated prediction of the visual stream, the red box indicates coordinates for attention predicted by the proprioceptive LSTM; (\textbf{E}) the ground truth target for comparison; (\textbf{F}) difference between the predicted visual stream and the ground truth target; (\textbf{G}) normalized joint angle trajectory predicted by the model compared with the corresponding ground truth; and (\textbf{H}) mean difference between the predicted joint angles and the ground truth.}
\label{vis_example}
\end{figure*}

An example of successful generation of a goal-directed action plan to achieve a linguistically represented goal, when trained with Group A1, is shown in Figure 2. A visuo-proprioceptive error of 0.0113 was observed in this successful example. This figure shows mental simulation of the generated motor plan and the expected visual trajectory associated with the linguistically specified goal (\textit{"put green on blue ."}). Figure 2(G) compares ground truth joint-angle trajectories of the test sequence with inferred trajectories from the planning process. Trajectories 0-4 (red, blue, green, yellow, purple) represent joint angles of all 5 active rotary joints of the robot arm and joint. Number 5 (brown) refers to linear actuators of the robot gripper. The visual stream shows every 5th time step of the generated sequence of the model. The current focal area, in terms of size and position of the attention transformation is indicated by a red square. Parameterization of the attention transformer is generated as an additional output of the multi-layer proprioceptive LSTM, as previously mentioned. Attention (red box in Figure 2(D)) is directed toward the object to be manipulated and the gripper when the gripper starts to approach the object. Contents of VWM-2 and VWM-1 are illustrated in Figure 2(A) and (B), respectively. The shape and color of the manipulated object are represented in VWM-2. Note that when the object is being moved it disappears from VWM-1 and appears in VWM-2. This information flow between VWM-1 and VWM-2 emerged in the visual network through training, which is essential for generalization to novel situations, based upon our previous work \cite{jeff2021}. We leverage this emergent mechanism in the current model to facilitate grounding of language to visuo-proprioceptive behavior. These observations show that a mental image of a continuous visuo-proprioceptive pattern can be generated using goal-directed planning to achieve a linguistically specified goal. 

It is evident that the model is adept at associating linguistically represented goals with corresponding visuo-proprioceptive sequences. Details of the model's performance for generalizing to unlearned object positions (U-P) with learned linguistically represented goals and to unlearned compositions (U-C) of linguistically represented goals are provided in Table S1. Despite fewer variations of linguistic composition in training, the model still maintained a low error $\leq {0.0405}$ (Group D3), for U-P. Figure 3 shows the comparison of generalization performance between U-P and U-C for different groups with the highest training ratio. Although U-P performance does not change significantly, depending on the composition scale, U-C performance improves as the variation of task composition in the training increases from Group D1 to Group A1.  

\begin{figure}
\centering
\includegraphics[width=0.99\textwidth]{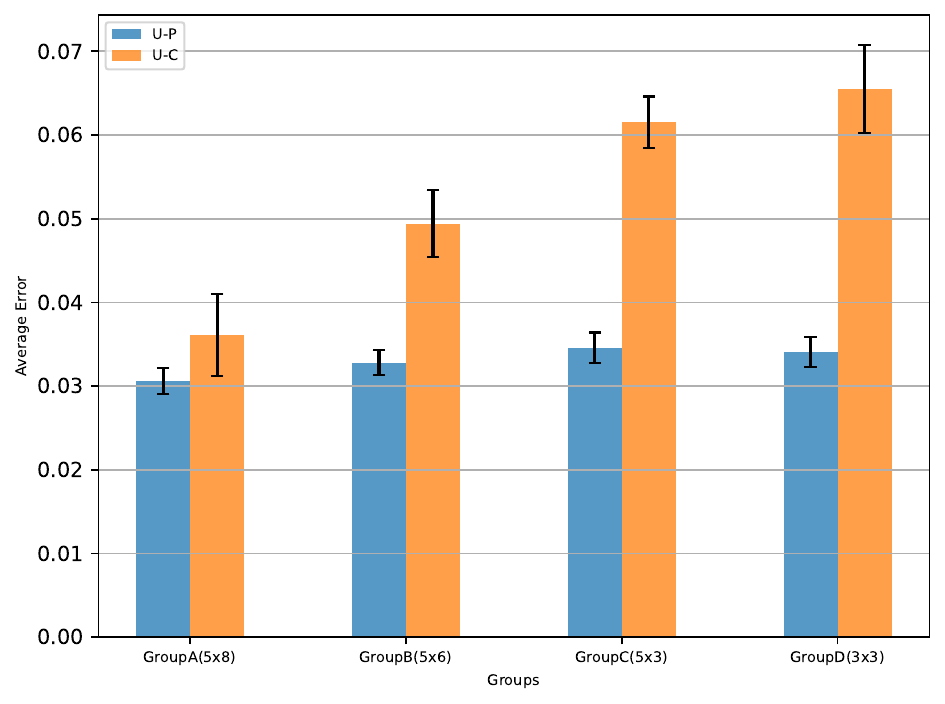}
\caption{Comparison of average visuo-proprioceptive error, for inference of visuo-proprioceptive plans, between unlearned object positions (U-P) and unlearned compositions (U-C) among groups with different number of compositions with the highest training ratio of 80$\%$.}
\label{group1_average_error}
\end{figure}

Figure 4 illustrates the difference in U-C performance between groups when trained with different training ratios, as mentioned above. In the majority of failed cases, the model confused colors of the object or misinterpreted the action to be performed on the object. This indicated that failure in generalization occurred at the abstract level, since low-level predictions still generated sequences corresponding to a different action or performed the specified action on a different object. Examples of visuo-proprioceptive sequence generated by the model in cases of successful and failed generalizations are shown in Figures S6 and S7, respectively. As just noted that even when the model fails to generalize, it still generates visuo-proprioceptive sequences that do not result in large errors. This likely explains why, despite poor performance in some instances, the average error remains relatively low.

\begin{figure}
\centering
\includegraphics[width=0.99\textwidth]{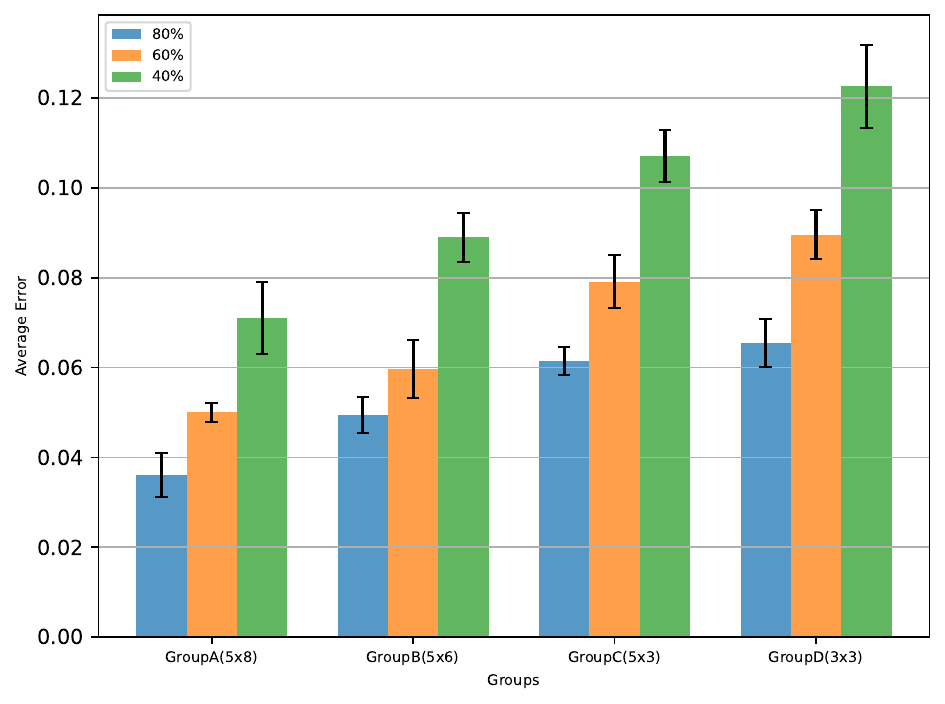}
\caption{Comparison of average visuo-proprioceptive  error, between groups with different training ratios for inference of visuo-proprioceptive plans to achieve unlearned compositions (U-C) of linguistically represented goals.}
\label{vp_performance}
\end{figure}

\begin{figure}[tbph]
\centering
\includegraphics[width=1\textwidth]{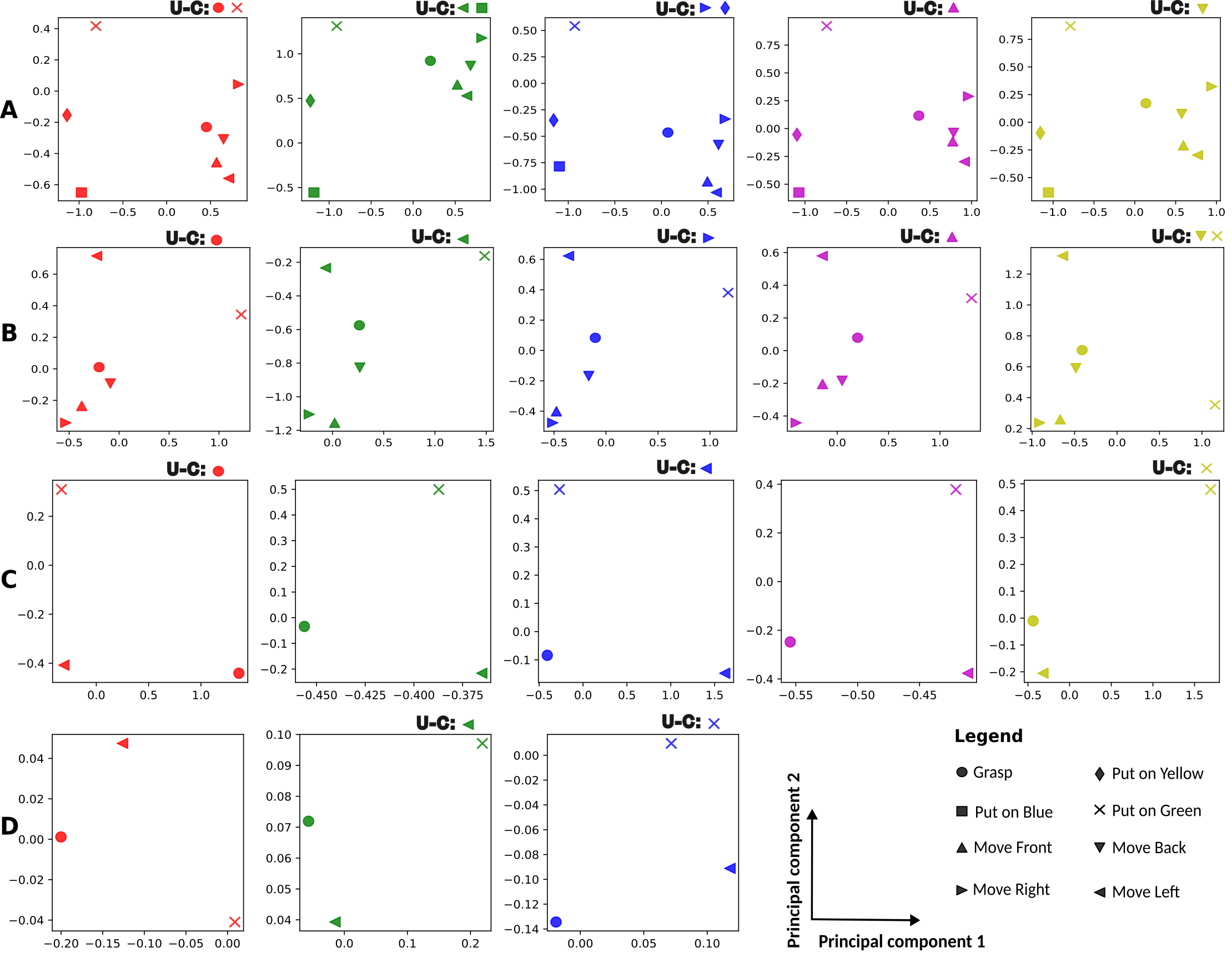}
\caption{Scatterplot of mean Kernel PCA values of latent state $\mathbf{PB}$ vectors for all groups with the highest training ratio. (\textbf{A}): Group A1 (5x8, 80\%) (\textbf{B}): Group B1 (5x6, 80\%) (\textbf{C}): Group C1 (5x3, 80\%) (\textbf{D}): Group D1 (3x3, 77\%). U-C refers to unlearned compositions that were used for testing. Colors of markers indicate the color of the object being manipulated. The variance explained by the two components of KPCA for all groups was greater than 90\%}
\label{kpca_pb_80_mean}
\end{figure}

We qualitatively analyzed a latent space, the parametric bias $\mathbf{PB}$ space, of linguistic-LSTM with kernel principal component analysis (KPCA) \cite{kpca}, using linear kernels. 
Figure 5 shows a scatterplot of mean KPCA values of each cluster in the $\mathbf{PB}$ space of the model for all groups with the highest training ratio. Figure S9 shows the data distribution corresponding to mean values shown here. The topology of hidden states corresponding to verb phrases (\textit{"grasp"}, \textit{"move left"}, etc) seems to have a common structure, when visualized separately for each object noun. For example, in Figure 5(A), representations of actions related to the stacking task are aligned on the left side and follow the order of "put X on green", "put X on yellow" and "put X on blue" (X refers to the object noun that corresponds to the object color in our experiments), from top to bottom. Similarly in Figure 5(A) the representation of actions related to \textit{"move"} are clustered on the right side and follow the order "move X right", "move X back", "move X front" and "move X left", from top to bottom. Also, the representation of actions related to \textit{"grasp"} is always between moving and stacking actions. Two things can be said from these observations. First, PB vectors corresponding to similar action categories become similar and second, this structure is largely similar for different colors, which implies compositionality between verbs and nouns.

This structure is more consistent among different object colors when the variety of task compositions in learning is high, as seen in Figures 5(A) and (B). In contrast, in Figures 5(C) and (D) topologies of hidden states are inconsistent among different object colors. We find that structural relationships between learned compositions are extrapolated by the model to generalize to unlearned compositions of nouns and verbs, illustrated by positions of unlearned compositions (U-C) (Figure 5). The model's ability to extrapolate this structure to unlearned compositions improves as the number of variations in task compositions increases. 
We also performed Welch's unequal variances T-test \cite{welch-t-test} to compare the generalization performance (for U-C) between groups, results of which are shown in Tables S3 and S4. The model showed robust generalization performance when trained with only 60\% of training data, for groups with a greater variety of task compositions (Group A and B). Groups trained with 40\% sparsity performed poorly, irrespective of the variety of task compositions in training. We observed a significant difference in model performance between subsequent groups, where groups with more variations in task compositions always outperformed groups with fewer elements for compositions. 

\subsubsection*{Language Inference from Observed Visuo-Proprioceptive Sequences}
The model’s ability to infer linguistically represented goals from visuo-proprioceptive behaviors was evaluated, with success defined by accurate inference of unlearned word sequences. Generalization performance for unlearned object positions (U-P) and linguistic compositions (U-C) is shown in Table S2, with Welch’s T-test results in Tables S5 and S6. Comparisons across different training ratios are illustrated in Figures S8, demonstrating that increased training composition size and ratios improve generalization. This is analogous to results obtained for inference of visuo-proprioceptive plan sequences (Figure 4). Although the model minimizes the error between generated visuo-proprioceptive sequences and the ground truth, success is measured by accuracy of inferred linguistic goals (Figure S8 and Table S2). Generalization performance in inferring appropriate linguistic goals is poorer than expected. This is possibly caused by the relatively noisy visual image sequence that was used for inference (Note that this goal inference, as well as training of the network, was performed by sampling real visual data operated by a physical robot.)

\subsection*{Experiment II: Ablation Study}
In order to assess the impact of visual attention and working memory on the current model's performance, we conducted ablation studies to evaluate the model's ability to generate mental simulation of visuo-proprioceptive sequences when provided with linguistically represented goals for Group A1 (5x8, 80\%). We ablated the visual attention and visual working memories (VWM-1 and VWM-2) (Table 2) and trained the model from scratch before evaluation. Ablation of either working memory or the attention module significantly affected the ability of the model to generate successful visual predictions, which resulted in poor generalization performance (Table 2). The model showed significantly degraded generalization performance for both unlearned object positions (U-P) and unlearned compositions (U-C). The proprioceptive prediction capability of the model was also reduced significantly, but less than visual prediction capability, due to ablation of the visual network. This is reflected in the proprioceptive accuracy (Table 2).

\begin{table}[h]
    \centering
    \caption{Ablation Study: Prediction Error}
    \resizebox{\textwidth}{!}{
    \begin{tabular}{c cc cc}
        \hline
            & \multicolumn{2}{c}{\bf{Visual Error ($\mu \pm{SD}$) \%}}  & \multicolumn{2}{c}{\bf{Proprioceptive Error ($\mu \pm{SD}$) \%}}\\
        \hline
            \bf{Condition} &\bf{U-P}  &\bf{U-C} &\bf{U-P}  &\bf{U-C} \\
        \hline
            \bf{VWM-1 \& 2 with attention}   &$0.0196\pm{0.0016}$   &$0.0249\pm{0.0038}$  &$0.0110\pm{0.0014}$    &$0.0117\pm{0.0020}$     \\ 
            \bf{only VWM-1 with attention}     &$0.0407\pm{0.0041}$   &$0.0543\pm{0.0011}$  &$0.0136\pm{0.0008}$    &$0.0158\pm{0.0014}$   \\ 
            \bf{only VWM-2 with attention}    &$0.0391\pm{0.0031}$   &$0.0506\pm{0.0074}$  &$0.0149\pm{0.0002}$    &$0.0155\pm{0.0015}$   \\
            \bf{VWM-1 \& 2 with no attention}   &$0.0480\pm{0.0050}$   &$0.0657\pm{0.0046}$  &$0.0154\pm{0.0007}$    &$0.0169\pm{0.0075}$ \\
            \bf{no VWM-1 \& 2 and no attention}   &$0.0734\pm{0.0052}$   &$0.0960\pm{0.0068}$ &$0.0198\pm{0.0004}$    &$0.0238\pm{0.0037}$  \\
        \hline
    \end{tabular}}
    \label{ablation_v}
\end{table}

The model performed comparatively better when there was at least one visual working memory with attention, compared with having no attention module. These results highlight the significance of the interaction between visual attention and working memory for generating accurate visuo-proprioceptive predictions. Further implications of these results are discussed below. 

\section*{Discussion}
This study investigated how generalization in compositionality can be achieved through the process of associative learning between action and language, even with limited amount of the experience, even with limited training data. We hypothesized that increasing task composition variation used in learning would improve generalization. This hypothesis was evaluated by conducting a set of simulated robotic experiments using the framework of active inference \cite{friston2010}. More specifically, we studied how robots can learn to generate goal-directed action plans by adequately inferring visuo-proprioceptive sequences to achieve linguistically specified goals. We also examined how such a robot can infer linguistically represented goals from observation of provided visuo-proprioceptive sequences. For this purpose, we built upon our previous work \cite{jeff2021} by adding a language processing LSTM and a PV-RNN in the associative layer, and by using a teleological approach for goal-directed planning \cite{Matsumoto2022}. The proposed model utilized a complex visual network with sub-modules, including visual attention and visual working memory modules (VWMs), to facilitate grounding of language to visuo-proprioceptive behavioral sequences. 

Our analysis of simulation experiment results to infer action plans for linguistically specified goals led to the following findings. First, generalization performance in learning unlearned compositions improves with increased vocabulary, as well as the training ratio. This is supported by an analysis of the $\mathbf{PB}$-space which showed more consistent relational structures among different concepts combining actions and object nouns. These emerge for cases with more variations of task composition in learning. 
Notably, we find representations in the $\mathbf{PB}$-space self-organized based on the similarity of actions. This validates our basic hypothesis. Second, performance for position generalization does not depend on the size of composition used in learning. This result can be understood by considering that position generalization competency should be developed in the lower level of the network model, which does not interact directly with linguistic compositional processing. 

Although several studies have investigated grounding of language with visuo-proprioceptive behavior using recurrent neural network models \cite{Sugita2005,Cangelosi2010,Cangelosi2018,Heinrich2018,akakzia2021decstr, taniguchi2023collective}, few have addressed the mechanism underlying compositionality. To the best of our knowledge, there have been no models that examined the compositional nature of language grounded in visuo-proprioceptive behavior, using the active inference framework. The current study explored
a possible underlying mechanism for linguistic compositionality, co-developed with sensory-motor skills to manipulate objects, using active inference. Extended studies should investigate how this mechanism can be developed gradually through incremental learning, as human children do.

By conducting ablation studies, we find that generalization performance in learning is significantly reduced when either visual working memory or visual attention is deleted from the model. This can be explained by considering that in the current model, visual image is perceived based on structures rather than as simple pixel patterns, by means of visual attention and working memory mechanisms. Actually, previous work \cite{jeff2021} using a similar visual network showed that coupled mechanisms of visual attention and working memory enable a manipulated object to be segmented from the background. Therefore, it is highly likely that marriage of structural visual information processing and compositional linguistic information processing enhances generalization in learning in the current task. Analysis of the model's performance in inferring linguistically represented goals from observation of visuo-proprioceptive sequences showed results analogous to those obtained for inferring visuo-proprioceptive plan sequences from provided linguistic goals. 


According to Hupkes et al. \cite{hupkes2020} compositionality of a neural network model should satisfy the following characteristics: \textit{systematicity} - it should systematically combine known parts and rules ; \textit{productivity} - it should extend its predictions beyond the lengths observed during training; \textit{substitutivity} - its predictions should be robust to synonymous substitutions; \textit{localism} - the degree to which the meaning of compositional expression depends on its immediate, local structures vs global structures; and \textit{overgenralization} - it should not favor particular rules or exceptions during training, i.e., it should not overgeneralize. We see evidence that our model demonstrates at least a rudimentary level of systematicity through its ability to generate appropriate visuo-proprioceptive sequence plans for unlearned combinations of actions and nouns, effectively generalizing to novel scenarios. This capability confirms that the model can synthesize new, meaningful sequences from previously learned parts and rules, meeting a key criterion for compositionality. Although testing for every aspect of compositionality is beyond the scope of the current study, it is a promising avenue for future research. 

We showed that generalization performance in learning unseen compositions increases as the size of the vocabulary and variations in task composition increase, evidenced by the best performance seen in the largest group (5x8 composition). This result offers a minimal potential solution to the poverty of stimulus problem \cite{chomsky2012poverty}. If the dimensions of composition increase to include not only verbs and object nouns, but also various modifiers such as adverbs and adjectives, and each dimension consists of hundreds of elements, as we experience in daily life, covering all possible combinations across all dimensions would result in a combinatorial explosion. Faced with this problem, our expectation is that the required amount of experience for generalization in learning is not proportional to the product of the number of elements across all dimensions, but rather is proportional to their summation, provided that the elemental size of each dimension is relatively large. Future studies should evaluate this possibility, which aims beyond generalization shown by typical neural network models, by conducting the same experiments under drastically scaled settings. The current study was necessary in order to examine basic mechanisms accounting for how generalization can be achieved in language-behavior compositionality with rigorous analysis, before scaling up the system. Additionally, it aimed to investigate how the model's capability improves with increments in sizes of individual elements in each dimension. 

Previous work with PV-RNN-based models \cite{wataru2020,Matsumoto2022} utilized the "online error regression" scheme, where prediction error serves as an input feature to retroactively correct the history of predictions (postdiction) while simultaneously predicting future behavior. This approach, while effective, is computationally intensive and relies on low-dimensional input features for feasibility. This limitation is especially significant given that our model evaluations were not conducted on physical robots. Even though the proposed model shows competitive performance in generating appropriate visuo-proprioceptive trajectories, compared to ground truth trajectories, executing the generated trajectories with a real robot may not yield the desired level of performance. Inferring object positions accurately from 64x64 RGB images in the video can result in large errors in the real world. A single-pixel error from noise in the 64x64 RGB image can result in a position error of several centimeters, which will significantly affect the behavioral performance of the robot, especially when the robot attempts to grasp objects. If a 256x256 RGB image can be used, the position error could be reduced to $<$1 cm.
This scheme, however, prohibits real time computation (It  takes several minutes to generate a single visuo-motor plan trajectory), since expensive back-propagation through time (BPTT) computation should be conducted through the convLSTM. Future studies should investigate more efficient solutions to speed-up this part of the computation, e,g., developing a C++ compiler for the whole system, instead of using the current Python-based program to achieve real-time operation of physical robots using the proposed model. Upon solving these issues the model will have the potential to be scaled up to more complex tasks with rich linguistic descriptions for cognitive robots to interact with the real world. We are actively working on this problem to make future iterations of the model more computationally efficient. 

Models like CLIP (Contrastive Language–Image Pre-training) \cite{radford2021} and CLIP-Guided Generative Latent Space Search (CLIP-GLaSS) \cite{Galatolo2021GeneratingIF} learn to associate images and textual descriptions in a joint embedding space. They employ a contrastive learning objective to align embeddings of images and their corresponding textual descriptions. While our model may share some features with these approaches, the $\mathbf{PB}$-vector is not a shared embedding for behavior and language. Instead it acts as a bottleneck to constrain both visuo-proprioceptive sequences and word sequences such that they share similar structures. Our rationale for not using a shared embedding space for visuo-proprioceptive behavior and language is to maintain flexibility in learning behavioral patterns that can achieve the same linguistic goals. For example, the robot may learn multiple trajectories to achieve the linguistically represented goal of \textit{"put red on green."}, depending on object positions.

A major limitation of the current study is the absence of communication in a societal context. Tasks were executed by a single robotic arm, lacking active engagement with other agents, which is essential for language development in humans \cite{tomasello2006social}. The model operates in a small workspace with a limited vocabulary. One possible way to scale the model is to increase the number of compositional elements (adverbs, adjectives, conjunctions, etc) to form longer sentences, as discussed previously. Moreover, extending the model to incorporate multiple agents, such as in RT-2\cite{lynch2022interactive}), each with their own models, could facilitate examination of communication dynamics within a societal context. 

Recent advances in LLMs have shown incredible performance with robots working in real-world environments \cite{flamingo_NEURIPS2022, ahn2022i, driess2023palme, brohan2023rt2}. These models, equipped with sophisticated language processing capabilities, have enabled robots to comprehend and generate human-like language, facilitating seamless interaction with users and enhancing their overall functionality. However, it's important to note a significant difference in the way language is acquired by these models compared to humans. While humans develop language skills through interaction with their environment, as we have tried to emulate in our model by intermingling linguistic cues with physical experiences and sensory inputs, the language capabilities of LLMs are predominantly acquired through passive exposure to vast linguistic datasets. The extent to which LLMs can truly understand language in a human-like manner \cite{chalmers1997conscious, marcus_2020, pezzulo-friston-llm,yoshida2023textmotiongroundinggpt4} is an intriguing question. As robotics continues to advance, bridging the gap between language understanding in machines and humans remains a key research challenge. Efforts to incorporate embodied interaction and sensorimotor experiences into language learning processes for robots hold promise to enhance the naturalness and robustness of their linguistic abilities in real-world scenarios.

\section*{Materials and Methods}


We propose a hierarchically organized generative model to handle multiple modalities including vision, proprioception and language. The model is trained end-to-end through supervised learning, utilizing training examples containing visuo-proprioceptive sequences paired with corresponding linguistic expressions. The model learns to generate visuo-proprioceptive sequences corresponding to associated linguistic expressions, by minimizing evidence free energy. The trained model is used to generate goal-directed plans in which the goal is represented by linguistic expression. Goal-directed visuo-proprioceptive sequences are generated by minimizing expected free energy, by means of active inference. Notably, this model inherits the vision module, with multiple visual working memory modules and visual attention \cite{jeff2021}. In order to integrate language into this model we employ the parametric bias ($\mathbf{PB}$) scheme proposed by Sugita and Tani \cite{Sugita2005}. For all experiments, the model was trained and evaluated with five random seeds, as described in the Implementation Details section of the Supplementary Materials. 

\subsection*{Model}
The overall architecture of the current model is illustrated in Figure 1(A). The model consists of RNN-based generative networks that handle prediction of vision, proprioception and language. These modalities are integrated by the associative network. There are three layers of stacked LSTM and convLSTM for the proprioception and vision networks, respectively. The language network is implemented as a single-layer LSTM with a parametric bias ($\mathbf{PB}$) vector (Figure 6). The associative network is a single-layer PV-RNN that connects to the top layer of the visual and proprioceptive networks. Language is bound to the associative network through a binding loss between the linguistic $\mathbf{PB}$ vectors and $\mathbf{\widetilde{PB}_t}$ that is generated by the associative network (Figure 6). 

Each layer in the visuo-proprioceptive pathway receives contextual information from neighboring layers. Top-down connectivity provides signal from the subsequent higher-level layer or from the associative layer of the model. Those layers propagate the prediction or belief of the network down to the sensorimotor level. A deconvolution operation is applied in the visual pathway, with dimensionality increasing from top to bottom. Visual and proprioceptive LSTM cells on the same layer of the model are connected via lateral connections. As in top-down processing, a deconvolution operation is applied to expand the low-dimensional space of proprioceptive representations to match the dimensions of the feature space of convLSTMs. Bottom-up connectivity sends neural activation from the lower layer of the model or the current sensory input, i.e. vision or proprioception, into the subsequent higher-level layer. Visual input is processed by an attention module, and a convolution operation is applied to reduce the dimension of projections to the next higher layer. Visual processing incorporates visual attention and saving/reading of visual images, using visual working memory (VWM). These are performed autonomously as part of the inference process.

The proprioceptive network predicts sequences of joint angles ($\widetilde{\mathbf{m}}_{t}$) and multiple low dimensional control signals ($\bm{\alpha}^{\text{att}}_{t}$ and $\bm{\alpha}^{\text{\VWMtwoSuffix}}_{t}$). These control signals act as parameters of visual attention and visual image transformation \cite{jeff2021, STN} in order to modulate information flow in the visual system. Visual attention performs dynamic adjustment of the pixel density in different regions of the image generated by the visual network. This allows the model to focus on and predict the visual appearance of manipulated objects in greater detail, while static parts of the generated images can be retrieved from the VWM-1. Furthermore, an additional parametric control of pixel-wise transformation \cite{STN} of images stored in VWM-2 allows the model to imitate dynamic changes of object images during its manipulation. This parametric control is performed by the output of proprioceptive LSTM ($\bm{\alpha}^{\text{\VWMtwoSuffix}}_{t}$). This transformation is limited to affine image transformations considering the nature of the task used in our experiments.

The visual network predicts pixel images of the currently attended region ($\mathbf{v}^{\text{att}}_{t}$) and a set of masks that are used to mix the predicted image with the contents of each VWM (see Figure 1 (A)). The final visual image is predicted through the interaction of this convLSTM prediction with parameterized visual image operations, including attention, inverse attention, fusion, and transformation. Attention is performed by application of the current attention filter, parameters of which are predicted by the proprioceptive LSTM on the plain visual image. Inverse attention is simply an inverse transformation of attended image. The fusion operations 
\begingroup
\mbox{(denoted by symbol \hspace{-0.7ex}\setbox0=\hbox{\includegraphics[width=1.0em]{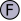}}
\parbox{\wd0}{\box0})}
\endgroup  fuses two sources of visual streams with a pixel-wise mixing ratio, using outputs and corresponding masks generated from the multi-layer convLSTM. Fusion operations are utilized to compose the final prediction, as well as to update the VWMs.

\begin{figure}[tbph]
\centering
\includegraphics[width=0.99\textwidth]{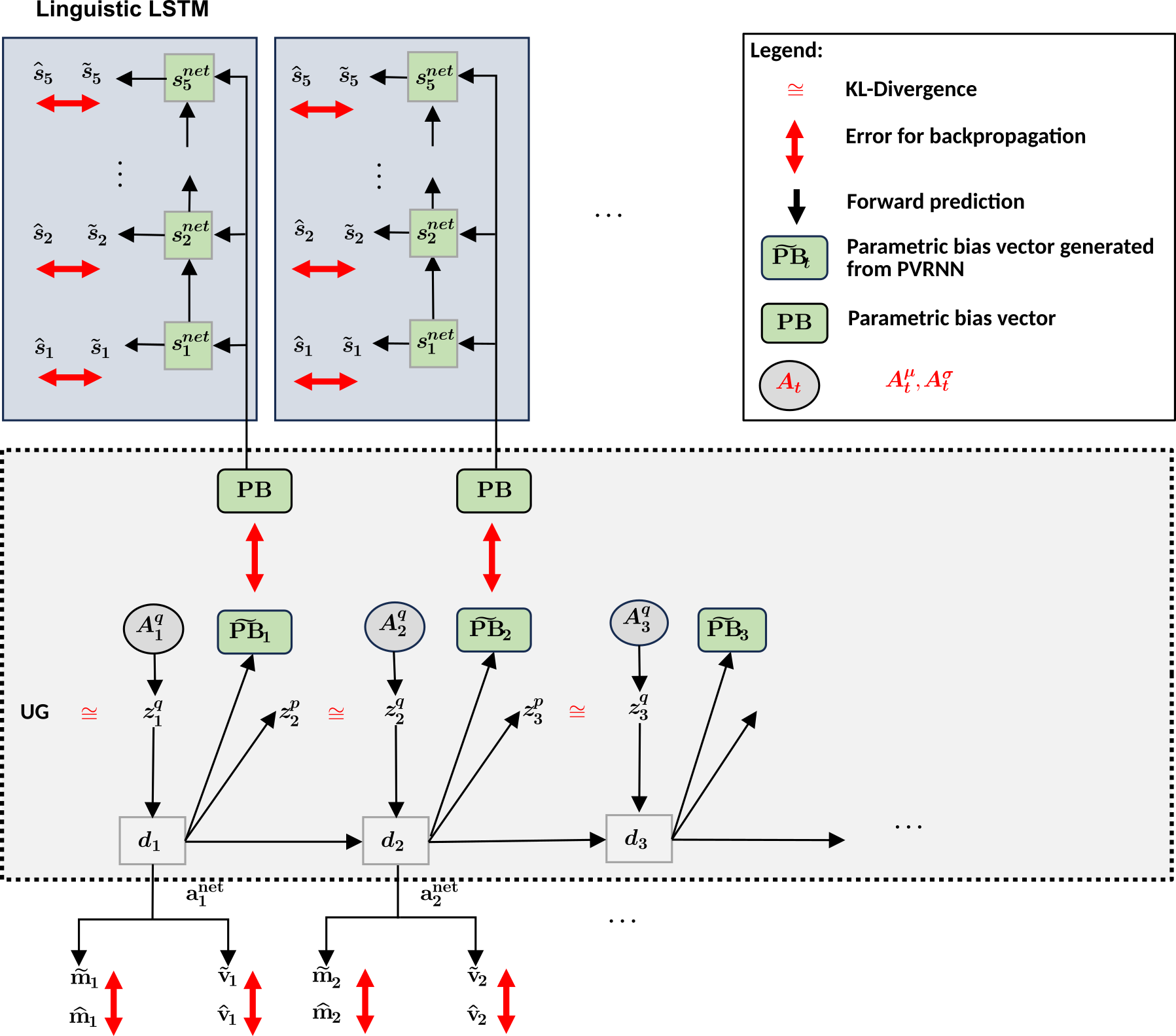}
\caption{Graphical representation of the associative PV-RNN and linguistic LSTM}
\label{language binding}
\end{figure}

Generation of top-down signals from the associative PV-RNN is based on the adaptive parameter $\mathbf{A}$  (representing the approximate posterior probability distribution in terms of the mean and standard deviation at each time step), which is optimized together with network parameters during training, to minimize the evidence free energy. The language network predicts a short sequence of words ($\widetilde{\mathbf{S}}$) at each time step of the visuo-proprioceptive sequence. Each word ($\widetilde{\mathbf{s}}_i$) is represented by a one hot vector, and the corpus is limited to 20 possible words. Language prediction depends on parameters of the linguistic-LSTM and the parametric bias ($\mathbf{PB}$) vector, which is constrained by the predicted $\widetilde{\mathbf{PB}}_t$ from the associative PV-RNN through a binding loss (equation 30) (Figure 6). 

During learning, updating adaptive variables and connectivity weights of the network is performed to minimize the reconstruction error between prediction and observation. To this end, back-propagation of the error is performed inversely through the aforementioned top-down and bottom-up pathways to update values of adaptive latent variables $\mathbf{A}$ and  $\mathbf{PB}$. In summary, the model represents a variational Bayes generative model in which learned multi-modal spatio-temporal sequence patterns, including proprioception, vision and language, can be reconstructed by adequately inferring the corresponding probabilistic latent variables $\mathbf{A}$ and $\mathbf{PB}s$. This is possible because all fabricated functions, such as visual attention, mask operation, and affine transformation, are designed as differentiable functions.

The lowest layers of vision, proprioception and language modules receive corresponding sensory inputs, pixel based images $\mathbf{v}_{t}$, the softmax representation of the current joint angle configuration $\mathbf{m}_{t}$, and a series of one-hot vectors as linguistic input $\mathbf{s}_{i}$. The computation in the input layers of the network can be defined as:
\begin{align}
    \mathbf{v}_{l=0, t}^{\text{net}} &= \ATT(\mathbf{v}_{t}, \bm{\alpha}^{\text{att}}_{t}),\\
    \mathbf{m}_{l=0, t}^{\text{net}} &= \text{SoftMax}(\mathbf{m}_{t}),\\
    \mathbf{s}_{l=0, i}^{\text{net}} &= \mathbf{s}_{i},
\end{align}

with visual attention transformation $\ATT(\mathbf{v}_{t}, \bm{\alpha}^{\text{att}}_{t})$ parameterized by $\bm{\alpha}^{\text{att}}_{t}$ applied to the visual input. We use the suffix ${i}$ to denote individual word steps of the language. It is important to note that while vision and proprioception are synchronized and share the same number of time steps, language is expressed as a sentence in which each word is represented by a one-hot vector. Therefore linguistic prediction, limited to five word steps, is predicted by the model at each step of the visuo-proprioceptive sequence (Figure 6).  

The proprioceptive prediction $\mathbf{m}^{\text{net}}_{t}$ as well as the low-dimensional parameterizations $\bm{\alpha}^{\text{att}}_{t}$ and $\bm{\alpha}^{\text{\VWMtwoSuffix}}_{t}$, which modulate attention and the affine transformation of VWM-2, are generated from the hidden states of the proprioceptive pathway as:

\begin{align}
    \mathbf{m}^{\text{net}}_{t} &= \FFN(\mathbf{m}^{\text{net}}_{l=1,t}), \label{eq:m_net}\\
    \bm{\alpha}^{\text{att}}_{t} &= \FFN(\mathbf{m}_{l=1, t}^{\text{net}}),\label{eq:alpha_att}\\
    \bm{\alpha}^{\text{\VWMtwoSuffix}}_{t} &= \FFN(\mathbf{m}_{l=1, t}^{\text{net}}),\label{eq:alpha_M2}
\end{align}
with $\FFN$ denoting a fully connected feed-forward network. Note that the $\FFN$s do not share the same connectivity weights.

Neural activation in the visual pathway (stacked convLSTM) for layer $l=1$ to $l=L$ at timestep $t$ is defined as: 
\begin{align}
    \mathbf{v}_{l, t}^{\textit{net}} &= 
    \begin{cases}
        \ConvLSTM\left(\mathbf{v}_{l-1, t}^{\text{net}}, \mathbf{m}_{l, t-1}^{\text{net}}, \mathbf{a}^{\text{net}}_{t-1}\right),& \text{if } l = L\\
        \ConvLSTM\left(\mathbf{v}_{l-1, t}^{\text{net}}, \mathbf{m}_{l, t-1}^{\text{net}}, \mathbf{v}_{l+1, t-1}^{\text{net}}\right),              & \text{otherwise.}
    \end{cases}
\end{align}

Neural activation in the proprioceptive pathway (stacked LSTM) is defined as:
\begin{align}
    \mathbf{m}_{l, t}^{\textit{net}} &= 
    \begin{cases}
        \LSTM\left(\mathbf{m}_{l-1, t}^{\text{net}}, \mathbf{v}_{l, t-1}^{\text{net}}, \mathbf{a}^{\text{net}}_{t-1}\right),& \text{if } l = L\\
        \LSTM\left(\mathbf{m}_{l-1, t}^{\text{net}}, \mathbf{v}_{l, t-1}^{\text{net}}, \mathbf{m}_{l+1, t-1}^{\text{net}}\right),              & \text{otherwise.}
    \end{cases}
\end{align}

The model uses only one LSTM layer with $\mathbf{PB}$ for the language network. Its neural activation is defined as:

\begin{equation}
    \mathbf{s}^{\text{net}}_{i} = \LSTM(\mathbf{s}^{\text{net}}_{i-1}, \mathbf{PB})
\end{equation}

As previously mentioned, the visual and proprioceptive pathways are connected by lateral connections in each layer of convLSTM and LSTM blocks, respectively. Additionally, the model includes an associative PV-RNN for a combined representation of both pathways in the highest layer. PV-RNN is comprised of deterministic $\mathbf{d}$ and stochastic $\mathbf{z}$ latent variables. The PV-RNN generates predictions from a prior distribution $\mathbf{p}$ and infers an approximate posterior distribution $\mathbf{q}$ by means of prediction error minimization on the generated sensory output $\mathbf{x}$. The prior generative model $\mathbf{p_\theta}$ is factorized as shown in following equation.
\begin{equation} \label{eq:genp}
\begin{split}
    & \mathbf{p_\theta}(\mathbf{x}_{1:T}, \mathbf{d}_{1:T}, \mathbf{z}_{1:T} | \mathbf{d}_0) = \\
    & \prod_{t=1}^T \mathbf{p_{\theta_x}}(\mathbf{x}_t | \mathbf{d}_t)  \mathbf{p_{\theta_d}}(\mathbf{d}_t | \mathbf{d}_{t-1}, \mathbf{z}_t) \mathbf{p}_{\theta_z}(\mathbf{z}_t | \mathbf{d}_{t-1}).
\end{split}
\end{equation}

The prior distribution $\mathbf{p_{\theta_z}(z_t} | \mathbf{d}_{t-1})$ is a Gaussian and it depends on $\mathbf{d}_{t-1}$, except at the initial time step $t=1$ which is fixed as a unit Gaussian with zero mean. Each sample of the prior distribution $\mathbf{z}_{t}^{p}$ is computed as shown in the following equation.
\begin{equation} \label{eq:p} 
\begin{aligned}
    \bm{\mu}^p_{t} &=
    \begin{cases}
        0, & \text{if } t=1 \\
        \text{tanh}(\bm{W}_{\mathbf{d},\mathbf{z}^{\bm{\mu}^p}}\mathbf{d}_{t-1}), & \text{otherwise}
    \end{cases}
    \\
    \bm{\sigma}^p_{t} &=
    \begin{cases}
        1, & \text{if } t=1 \\
        \exp(\bm{W}_{d,z^{\bm{\sigma}^p}}\mathbf{d}_{t-1}), & \text{otherwise}
    \end{cases}
    \\
    \mathbf{z}^p_{t} &= \bm{\mu}_{t}^p + \bm{\sigma}_{t}^p * \bm{\epsilon} .
\end{aligned}
\end{equation}

where $\bm{\epsilon}$ is a random noise sample such that $\bm{\epsilon} \sim \mathcal{N}(0, I)$. $\mathbf{W}$ is the connectivity weight matrix. We omit the bias term in all equations for the sake of brevity. 

Since computing the true posterior distribution is intractable, the model infers an approximate posterior ($\mathbf{q}_\varphi$) at time step $t$, $\mathbf{z}_t^q$ computed as shown in equation 12. $\mathbf{A}_{1:T}^\mu, \mathbf{A}_{1:T}^\sigma$ are adaptive variables, inferred through back-propagation by minimizing the prediction error and the complexity term, as detailed below. These are used to compute the mean and standard deviation for the approximate posterior at each step in a sequence.

\begin{equation} \label{eq:q} 
\begin{aligned}
    \bm{\mu}^q_{t} &= \text{tanh}(\mathbf{A}^\mu_{t}) ,\\
    \bm{\sigma}^q_{t} &= \exp(\mathbf{A}^\sigma_{t}) ,\\
    \mathbf{z}^q_{t} &= \bm{\mu}_{t}^q + \bm{\sigma}_{t}^q * \bm{\epsilon} .\\
\end{aligned}
\end{equation}

We use a multiple timescale recurrent neural network (MTRNN) \cite{Yamashita2013}, adapted for a single layer, as the RNN for the associative PV-RNN layer. The deterministic latent variable of the associative layer $\mathbf{a}^{\text{net}}_{t}$ is computed as follows, 

\begin{equation} \label{eq:hiddenlayer} 
\begin{aligned}
    \mathbf{d}_t = \left(1 - \frac{1}{\tau}\right)\mathbf{d}_{t-1}
    + \frac{1}{\tau} \biggl( \mathbf{W}_{a,a}\mathbf{a}^{\text{net}}_{t-1} + \mathbf{W}_{z,a}\mathbf{z}^{q}_{t} \\ + \mathbf{W}_{v,a}\mathbf{v}_{l=L, t-1}^{\text{net}} + \mathbf{W}_{m,a}\mathbf{m}_{l=L, t-1}^{\text{net}}\biggr), \\
\end{aligned}
\end{equation}

\begin{equation}
    \mathbf{a}^{\text{net}}_{t} = \tanh(\mathbf{d}_t).
\end{equation}

where $\tau$ is the time constant that determines the rate at which the network integrates information over time. The associative layer also predicts a parametric bias ($\mathbf{\widetilde{PB}_t}$) vector at each time step, which is bound to the $\mathbf{PB}$ vector of the language module through a binding loss (equation 30). The $\mathbf{\widetilde{PB}_t}$ vector is computed as:

\begin{equation}
    \mathbf{\widetilde{PB}_t} = \tanh(\mathbf{W}_{d,pb}\mathbf{d}_{t})
\end{equation}
$\mathbf{W}_{d,pb}$ is the connectivity weight matrix between $\mathbf{d}$ and $\textbf{PB}$. 

Our previous study \cite{jeff2021} showed that visual image transformations by attention and inverse attention are among the most important elements for successful development of visual working memory function during end-to-end learning. As mentioned previously, visual attention is performed by an attention transformation parameterized by scaling and coordinates of a focal position. These parameters are generated by the proprioceptive multi-layer LSTM, which receives top-down signals from the associative PV-RNN in the higher level. This means that optimal parameters for visual attention during training and goal-directed planning are determined by the inference of optimal latent-state values $\mathbf{A}_{1:T}$. 

The visual attention and visual working memory systems are applied to the output of the convLSTM block, which include prediction of the attended visual image $\mathbf{v}^{\text{net}}_{t}$ and a set of masks, computed as:
 
\begin{align}
    \mathbf{v}^{\text{net}}_{t} &= \tanh(\deconv(\mathbf{v}_{l=1, t}^{\text{net}})),\\
    \begin{bmatrix}
           \mathbf{g}^{\text{\VWMoneSuffix}}_{t} \\
            \mathbf{g}^{\text{pred}}_{t} \\
    \end{bmatrix} &= \ATT^{-1}(\textit{Sig}(\deconv(\mathbf{v}_{l=1, t}^{\text{net}})), \bm{\alpha}^{\text{att}}_{t})\\
    \begin{bmatrix}
           \mathbf{g}^{\text{\VWMtwoSuffix}}_{t} \\
           \mathbf{g}^{\text{net}}_{t}
    \end{bmatrix} &= \textit{Sig}(\deconv(\mathbf{v}_{l=1,t}^{\text{net}})),
\end{align}

with a sigmoidal activation function $\textit{Sig}$. Note that the $\deconv$ operations do not share the same connectivity weights. The masks $\mathbf{g}^{\text{\VWMoneSuffix}}_{t}$ and $\mathbf{g}^{\text{\VWMtwoSuffix}}_{t}$ modulate the pixel-wise update of the VWM-1 and VWM-2, respectively. Furthermore, the masks $\mathbf{g}^{\text{pred}}_{t}$ and $\mathbf{g}^{\text{net}}_{t}$ decide how much the final visual prediction depends on the VWMs or $\mathbf{v}^{\text{net}}_{t}$ (see figure 1(A)).  

Details of network-wise operations for visual working memories, VWM-1 and VWM-2, are described by the following equations:

\begin{equation}\label{eq:bufl0} 
\begin{split}
    \mathbf{vwm}^{\text{\VWMoneSuffix}}_{t+1} = (1-\mathbf{g}^{\text{\VWMoneSuffix}}_{t})\odot \mathbf{vwm}^{\text{\VWMoneSuffix}}_{t}\\ + \mathbf{g}^{\text{\VWMoneSuffix}}_{t}\odot
    \ATT^{-1}(\mathbf{v}^{\text{att}}_{t},\bm{\alpha}^{\text{att}}_{t}).
\end{split}
\end{equation}
Equation 19 describes how contents of VWM-1 ($\mathbf{vwm}^{\text{\VWMoneSuffix}}_{t+1}$), can be updated, where $\mathbf{g}^{\text{\VWMoneSuffix}}_{t}$ denotes a pixel-wise mask and $\ATT^{-1}$ performs inverse attention transformation, parameterized by $\bm{\alpha}^{\text{att}}_{t}$ (equation 5), on the predicted attended visual image $\mathbf{v}^{\text{att}}_{t}$. The element-wise multiplication operator denoted by symbol $\odot$ fuses the visual stream and mask.

\begin{equation}\label{eq:bufl1}
\begin{split}
    \mathbf{vwm}^{\text{\VWMtwoSuffix}}_{t+1} =\;& \mathbf{g}^{\text{\VWMtwoSuffix}}_{t}\odot\TRAN(\mathbf{vwm}^{\text{\VWMtwoSuffix}}_{t}, \bm{\alpha}^{\text{\VWMtwoSuffix}}_{t})\\ &+ (1-\mathbf{g}^{\text{\VWMtwoSuffix}}_{t})\odot \mathbf{v}^{\text{att}}_{t}.
\end{split}
\end{equation}

Equation 20 describes how VWM-2, $\mathbf{vwm}^{\text{\VWMtwoSuffix}}$, can be updated. The variable $\mathbf{g}^{\text{\VWMtwoSuffix}}_{t}$ denotes a pixel-wise mask that defines the fusion of transformed contents $\TRAN(\mathbf{vwm}^{\text{\VWMtwoSuffix}}_{t} , \bm{\alpha}^{\text{\VWMtwoSuffix}}_{t})$ of VWM-2  with $\mathbf{v}^{\text{att}}_{t}$, the predicted image in the attended feature space from the previous step. This ensures that the contents of  $\mathbf{vwm}^{\text{\VWMtwoSuffix}}$ are influenced only by the visual prediction in the attended region and by the transformed $\mathbf{vwm}^{\text{\VWMtwoSuffix}}$. Prediction $\mathbf{v}^{\text{att}}_{t}$ of attended visual images is performed by a fusion of the predictions made by the convLSTM, $\mathbf{g}^{\text{net}}_{t}$ and the contents of VWM-2, defined by:

\begin{equation}
    \mathbf{v}^{\text{att}}_{t}=\, 
    \mathbf{g}^{\text{net}}_{t}\odot
    \mathbf{v}^{\text{net}}_{t}+
    (1-\mathbf{g}^{\text{net}}_{t})\odot
    \TRAN(\mathbf{vwm}^{\text{\VWMtwoSuffix}}_{t},\bm{\alpha}^{\text{\VWMtwoSuffix}}_{t}).  
\end{equation}

The visual output $\mathbf{\widetilde{v}}_{t}$ is computed by fusion of the contents of VWM-1, $\mathbf{vwm}^{\text{\VWMoneSuffix}}_{t}$, with the predicted attended image $\mathbf{v}^{\text{att}}_{t}$. Inverse attention transformation $\ATT^{-1}$ is applied to $\mathbf{v}^{\text{att}}_{t}$ to make it possible to fuse with the contents VWM-1, defined as:

\begin{equation}
    \mathbf{\widetilde{v}}_{t}=\,
    \mathbf{g}^{\textit{pred}}_{t}\odot
    \ATT^{-1}(\mathbf{v}^{\text{att}}_{t},\bm{\alpha}^{\text{att}}_{t}) +
    (1-\mathbf{g}^{\text{pred}}_{t})\odot
    \mathbf{vwm}^{\text{\VWMoneSuffix}}_{t}.
\end{equation}

The final proprioceptive prediction is generated by a decoding of the softmax encoded predictions of the LSTM to get the trajectory, $\mathbf{\widetilde{m}_{t}}$, of joint angle configurations:
\begin{equation}
   \mathbf{\widetilde{m}_{t}} = \text{SoftMax}^{-1}(\mathbf{m}^{\text{net}}_{t}).
\end{equation}

The final linguistic prediction, $\mathbf{\widetilde{s}}_{i}$, is computed through a fully connected output layer followed by a softmax activation function to get a one-hot vector representation for each word, 
\begin{equation}
   \mathbf{\widetilde{s}}_{i} = \text{SoftMax}(\FFN(\mathbf{s}^{\text{net}}_{i}))
\end{equation}

The sentence predicted at every time step of the behavior is the same and is defined as $\mathbf{\widetilde{S}} = (\mathbf{\widetilde{s}_1}, \mathbf{\widetilde{s}_2}, \mathbf{\widetilde{s}_3}, \mathbf{\widetilde{s}_4}, \mathbf{\widetilde{s}_5})$. Examples of sentences describing actions performed by the robot are \textit{"put red on green ."}, \textit{"grasp red ."}, \textit{"move red left ."}, \textit{etc}, where each word is represented by one hot vector. Note that the sentences are of different lengths; therefore, in order for all sentences to be the same length, the remaining steps are masked with zero vectors to get a maximum of five word steps. The final character, $\mathbf{\widetilde{s}_5}$ of every sentence is always the vector corresponding to \textit{"."}, indicating the end of the sentence. 

\subsection*{Learning by Minimizing Free Energy}

The model learns to generate visuo-proprioceptive sequences by minimizing the evidence free energy $\mathcal{F}$ \cite{FristonFEP}. Free energy is comprised of an accuracy term and a complexity term, as explained in the Materials and Methods  of the Supplementary Materials (subsection entitled "Training by minimizing Free Energy"). In the current model, during learning the free energy is minimized by updating model parameters that include network weights, biases, the parametric bias $\mathbf{PB}$, and the adaptive latent variable $\mathbf{A}_{t:T}$. The accuracy term can be defined as the sum of reconstruction errors from each modality. These are defined as: 
\begin{align} 
    L_{\mathbf{v}} &= \sum_{t=1}^{\textit{T}}L_{\mathbf{v},t}=\sum_{t=1}^{\textit{T}}\mathbf{c}^{\text{att}}_{t}\odot(\mathbf{\widehat{v}}_{t} - \mathbf{\widetilde{v}}_t)^2, \quad\\
    L_{\mathbf{m}} &=\sum_{t=1}^{\textit{T}}L_{\mathbf{m},t}=\sum_{t=1}^{\textit{T}}\textit{D}_{\text{KL}} (\text{SoftMax}(\mathbf{\widehat{m}}_{t})||\mathbf{\widetilde{m}_{\text{net,t}}}),\\
    L_{\mathbf{S}} &= \sum_{i=1}^{5}(\mathbf{\widehat{s}_i} - \mathbf{\widetilde{s}_i})^2
\end{align}

$L_{\mathbf{v}}$ is the visual error, $L_{\mathbf{m}}$ is the proprioceptive error and $L_{\mathbf{S}}$ is the language error. $\mathbf{c}^{\text{att}}_{t}$ is error term that balances the contribution of the focal and peripheral regions of the visual error. This impedes over-representation of back-propagated gradients of the focal area. Details of calculation of $\mathbf{c}^{\text{att}}_{t}$ are described in section S1 of Supplementary Materials. Lengths of visuo-proprioceptive sequences are denoted by $\textit{T}$. The proprioceptive target for time step $t$ is a softmax encoding of joint angles $\mathbf{\widehat{m}_t}$. Visual targets at each time step are observed images $\mathbf{\widehat{v}}_t$ and language targets are a sequence of words $\mathbf{\widehat{s}}_i$. Note that language has a maximum of five words that describe each task. 

Additionally, we introduce a binding loss between ${\mathbf{\widetilde{PB}_t}}$ predicted and generated by PV-RNN and ${\mathbf{PB}}$ inferred by linguistic LSTM shown as:

\begin{equation}\label{eq:binding_loss} 
    L_{\mathbf{pb}} = \mathbf{k}*\sum_{t=1}^{\textit{T}}(\mathbf{\widetilde{PB}_t} - \mathbf{{PB}})^2    
\end{equation}
$\mathbf{k}$ is the binding coefficient that influences the strength of binding between language and behavior. The associative network must predict a constant parametric bias at all time steps in order to minimize this loss. This imposes a constraint on the associative network to generate visuo-proprioceptive predictions that are bound to language. This allows flexibility in representation, since the same linguistic expression can be associated with the task performed at random positions of objects in the workspace. 

The loss optimized by the model during training is then defined as: 

\begin{equation}\label{eq:trainingloss}
    L = L_{\mathbf{v}} + L_{\mathbf{m}} + L_{\mathbf{s}} + L_{\mathbf{pb}} + \mathbf{w} \sum_{t=1}^{\textit{T}}D_{\text{KL}}(q_{\varphi}(\mathbf{z}_t|X)||p_{\theta}(\mathbf{z}_t))
\end{equation}

${X}$ indicates the sensory observation in all modalities. The complexity term is the KL divergence between the approximate posterior and prior distributions. Note the formal similarity between equation 31 and equation 25 (and 32) where the accuracy corresponds to the various components of training loss, $L$. In the current model, free energy is modified by inclusion of the meta-prior $\mathbf{w}$ that weights the complexity term. $\mathbf{w}$ is a hyperparameter that affects the degree of regularization in PV-RNN, determined by the experimenter.

The model updates the adaptive latent variable $\mathbf{A}_{1:T}$ (used to compute $q_{\varphi}$ as described above) with network weights and biases in order to minimize the free energy, which consists of the prediction error between the predicted sensation and the observed ground truth and the complexity term, which is the KL-divergence between the prior ($p_{\theta}$) and the approximate posterior ($q_{\varphi}$). The prior is also learned through minimization of free energy. The learned prior is utilized to generate sensory predictions during the inference phase. Algorithm 1, in the Supplementary Materials, describes the learning process for a single training sequence.

\subsection*{Goal-directed planning using Active Inference}
Active inference proposes that action generation is a way to minimize error between the desired goal state and the predicted goal state by acting appropriately on the environment (see Supplementary Materials under the subsection "Linguistically Specified Goal-Directed Planning by Active Inference"). A standard interpretation of goal-directed planning is to assume that the goal is presented at the distal step of a behavioral sequence. In the current model, we use a teleological approach to goal-directed planning proposed by Matsumoto et.al \cite{Matsumoto2022}. The central idea is that goal expectation is generated at every time step instead of just at the distal step. In our model, the expected free energy or goal loss is defined as:
\begin{equation}\label{eq:planningloss_lang goal}
    L^g = L^g_{\mathbf{S}} + L^g_{\mathbf{pb}} +\mathbf{w} \sum_{t=1}^{\textit{T}}D_{\text{KL}}(q_{\varphi}(\mathbf{z}_t)||p_{\B{\theta}}(\mathbf{z}_t)) ,
\end{equation}
$L^g_{\mathbf{S}}$ is the reconstruction error between predicted language and goal language, $L^g_{\mathbf{PB}}$ is the binding loss between the corresponding parametric bias $\mathbf{PB}$ vectors. The KL divergence (complexity term) between the approximate posterior and prior distributions is weighted by the meta-prior $\mathbf{w}$. The length of the visuo-proprioceptive sequence is denoted as $\textit{T}$. Back-propagation of error is performed for inferring adaptive latent variables (A) in the associative PV-RNN and the parametric bias (PB) in the linguistic LSTM (while the connectivity weights of the whole network are fixed) by which, plans, in terms of visuo-proprioceptive sequences are generated. Algorithm 2, in the Supplementary Materials, describe both types of inference processes. The model performance is evaluated based on its ability to successfully generate action plans (visuo-proprioceptive sequences) when given a specific linguistically represented goal. 

\subsection*{Language Inference from Observed Visuo-Proprioceptive Sequences}
The model can also generate linguistic predictions that correspond to observed visuo-proprioceptive sequences, by optimizing approxiamte posterior ($q_{\varphi}$) and parametric bias (while the connectivity weights of the whole network are fixed) in order to minimize the free energy of the visuo-proprioceptive sequences (Figure S11). The free energy that is minimized for inference of linguistically represented goal is defined as:

\begin{equation}\label{eq:planningloss_lang inf}
    L^g = L^g_{\mathbf{v}} + L^g_{\mathbf{m}} + L^g_{\mathbf{pb}} +\mathbf{w} \sum_{t=1}^{\textit{T}}D_{\text{KL}}(q_{\varphi}(\mathbf{z}_t)||p_{\B{\theta}}(\mathbf{z}_t)) ,
\end{equation}
$L^g_{\mathbf{v}}$ and $L^g_{\mathbf{m}}$ is the error between observed ground truth and the vision and proprioceptive predictions, respectively. Algorithm 3, in the Supplementary Materials, describe both types of inference processes.

\bibliographystyle{Science}

\bibliography{references}

\newpage

\section*{Acknowledgments}
This work was supported by funding from Okinawa Institute of Science and Technology (OIST) Graduate University. JT was partially supported by the Japan Society for the Promotion of Science (JSPS) KAKENHI, Transformative Research Area (A): unified theory of prediction and action [24H02175]. The authors are grateful for the help and support provided by the lab members in the Cognitive Neurorobotics Research Unit and the Scientific Computing section of Research Support Division at OIST. We thank Takazumi Matsumoto for providing the base code for data collection and Steven Aird for editing the manuscript.

\newpage

\section*{Supplementary Materials}
Materials and Methods\\
Supplementary text\\
Figs. S1 to S11\\
Tables S1 to S6\\
Algorithms for Learning and Inference\\

\setcounter{figure}{0}
\setcounter{table}{0}
\makeatletter 
\renewcommand{\thefigure}{S\@arabic\c@figure}
\renewcommand{\thetable}{S\@arabic\c@table}
\makeatother

\subsection*{Materials and Methods}
\def\UrlBreaks{\do\/\do-\do\&\do.\do:}
The code used for experiments is available at: \url{https://github.com/oist-cnru/FEP-based-model-of-Embodied-Language.git}. Access to the data will also be made available through the repository.

\subsubsection*{Training Data Acquisition}

We collected RGB video data from a fixed camera and joint angle trajectories from the robotic arm (Torobo Arm; Tokyo Robotics Inc.), shown in Figure S1, with 7 degrees of freedom. Rotation of the end effector is not used in these experiments, so it is omitted. The actuator is programmed in the joint-space position control in order to perform different object manipulation tasks on objects placed at random positions in the work space. Trajectories generated from the programmed controller are recorded to train the model. Object positions are sampled from a workspace of 10x10 and 8x8 grid for training and testing data, respectively. Trajectories were recorded at 20 Hz and down-sampled to reduce computational costs. There was approximately a 10$\%$ variation in the sequence lengths of the collected data. Video data were down sampled to sequence of 64x64 RGB data. After pre-processing, the resulting visuo-proprioceptive sequences were 50 $\pm$ 3 time steps for all trajectories. The recorded dataset contains 400 sequences (10 sequences for each noun-verb combination), and with each sequence generated from different initial object positions. Volumes of training and testing data differ based on the number of compositional elements in each group. These are separated into four groups; Group A (5x8) is a permutation of 5 nouns and 8 verbs for a total of 40 combinations, Group B (5x6) is a permutation of 5 nouns and 5 verbs for a total of 30 combinations, Group C (5x3) is a permutation of 5 nouns and 3 verbs for a total of 15 combinations, and Group D (3x3) is a permutation of 3 nouns and 3 verbs for a total of 9 combinations. These combinations were trained with three degrees of sparsity to evaluate the robustness of the model.

\subsubsection*{Implementation Details} 
The associative network, which integrates the vision, proprioception, and language modules, contains a single PV-RNN layer with 256 neurons and 20 stochastic units $\mathbf{z^p}$. The posterior units $\mathbf{z}^q$ have the same dimensions as $\mathbf{z^p}$. The language network consists of a single layer of LSTM with 40 neurons and a 10 dimensional parametric bias. The proprioceptive and visual pathways of the model are based on three layers of multi-layer LSTM cells and convLSTM cells, respectively. The multi-layer convLSTM contain 16, 32, and 64 feature maps from the lowest to the highest layer. To project features to the next higher layer, a convolutional kernel of size 5x5, stride 2x2, and padding size of 2x2 were used. A deconvolutional kernel size of 6x6, stride 2x2, and padding of 2x2 were used to project features to the consequent lower layer. For lateral connections from hidden states of the proprioceptive pathway to the visual pathway and the projection from the associative LSTM, convolutional kernel sizes are selected in such way as to match the feature dimensionality of the corresponding layer of the visual pathway. Forward transformation of the attention transformer down-scales the resolution by a factor of $0.625$ (64x64 to 40x40 pixels), to fit the size of prerecorded datasets. The multi-layer LSTM  of the proprioceptive pathway contains 256, 128, and 64 neurons from the lowest to the highest layer. The proprioceptive pathway for prediction of proprioception and parameterization of the attention modules was performed by a multilayer perceptron (MLP) with one hidden layer of 256 neurons. Layer normalization and rectified linear unit (ReLU) activation functions are utilized. The model was trained and evaluated with five random seeds (0, 5, 10, 15 and 20) for all experimental evaluations. 

\subsubsection*{Training by minimizing Free Energy}
Training of the model was done by minimizing free energy. For a generative model $\mathbf{p}_{\mathbf{\theta}}(\mathbf{X})$, free energy is defined as: 

\begin{equation}\label{eq:vfe}
    \mathcal{F} = -\underbrace{E_{q_{\varphi}(\mathbf{z})}[\operatorname{ln}p_{\theta}(\mathbf{X|z})]}_{\text{a) Accuracy}} + \underbrace{D_{\mathrm{KL}} [q_{\varphi}(\mathbf{z|X})||p(\mathbf{z})]}_{\text{b) Complexity}}, 
\end{equation}

with latent variable $\mathbf{z}$, observations $\mathbf{X}$ and model parameters ${\varphi}$ and ${\theta}$. The accuracy term (a) is the expected likelihood of observed sensations $\mathbf{X}$, given $\mathbf{z}$. This can also be interpreted as the reconstruction error between predicted sensation and ground truth. The complexity term (b) facilitates regularization of the model by minimization of the Kullback-Leibler (KL) divergence between the approximate posterior $q_{\varphi}(\mathbf{z|X})$ and prior $p(\mathbf{z})$. The evidence free energy $\mathcal{F}$ can be minimized with respect to the posterior distribution $q_{\varphi}(\mathbf{z})$ as:

\begin{equation}\label{eq:vfeopt}
    q_{\mathbf{\varphi}}(z) = \underset{ q_{\mathbf{\varphi}}(z)}{{arg}\!\min}\,\mathcal{F}
\end{equation}
Negative free energy is also known as an evidence lower bound in machine learning literature\cite{Bishop2006}. Therefore, minimizing free energy is equivalent to maximizing the evidence or marginal likelihood for a generative model. In the current model, during learning the free energy is minimized by updating model parameters that include network weights, biases, the parametric bias $\mathbf{PB}$, and the adaptive latent variable $\mathbf{A}_{t:T}$.

Training of the model was done by minimizing the loss function in equation 31, performed using the ADAM optimizer \cite{kingma2017adam}. 
Optimization of parametric bias $\mathbf{PB}$, adaptive latent variables $\mathbf{A}$, weights, and biases was performed for over 5000 epochs, until convergence of learning. The learning rate was set to $5 \times 10^{-4}$ and the hyper-parameter $w$ was set to $0.05$. The binding coefficient $\mathbf{k}$ for the binding loss between predicted $\mathbf{\widetilde{PB}}$ and language parametric bias $\mathbf{PB}$ is set to 100. To prevent instability during training, i.e., \textit{exploding gradient problem}, we performed gradient clipping \cite{pascanu13}, which re-scales gradients based on the ${\ell}_2$-norm in case the norm of the gradients exceeds $0.2$. The mean and standard deviation of the prior at first time step was set to 0 and 1, respectively. 

\subsubsection*{Linguistically Specified Goal-Directed Planning by Active Inference}
The trained model is evaluated via active inference. More specifically, a sequence of action is inferred by minimizing the \emph{expected free energy} $\mathcal{G}$ for the future considering possible effects of actions $a$ applied to the environment. The expected free energy $\mathcal{G}$ is defined as:

\begin{equation}\label{eq:efe}
    \mathcal{G} = -\underbrace{E_{q_{\mathbf{\varphi}}(z)}[\operatorname{ln}p_{\mathbf{\theta}}(X(a)|z)]}_{\text{a) Accuracy in future}} + \underbrace{D_{\mathrm {KL} } [q_{\mathbf{\varphi}}(z)||p(z)]}_{\text{b) Complexity}}, 
\end{equation}

${X}$ is the preferred sensation given as a function of action $a$. The model performs goal-directed planning using active inference (Figure S10), by minimizing the expected free energy of a goal state represented by language, e.g. \textit{"put red on green ."}, that is associated with the corresponding visuo-proprioceptive sequences. The expected free energy $\mathcal{G}$ can be minimized with respect to the posterior distribution $q_{\varphi}(\mathbf{z})$ as:

\begin{equation}\label{eq:aiopt}
    q_{\mathbf{\varphi}}(z) = \underset{ q_{\mathbf{\varphi}}(z)}{{arg}\!\min}\,\mathcal{G}
\end{equation}

The trained model was evaluated for its ability to successfully generate goal-directed visuo-proprioceptive plans to achieve linguistically specified goals by means of active inference. Planning of action for previously unseen object position (U-P) as well as unlearned language composition (U-C) was done by minimizing the expected free energy defined in equation 30.  In order to update the posterior latent variables ($\mathbf{A}$) and parametric bias ($\mathbf{PB}$) at each epoch, the visuo-proprioceptive sequence was generated 5 times by repeated stochastic sampling of $\mathbf{A}$, the mean and variance of which were inferred. The values of $\mathbf{A}$ and linguistic $\mathbf{PB}$ that result in the smallest planning error ($L^g$) after 50 iterations of inference was selected as the final result of the planning process for the given goal. The visuo-proprioceptive sequences generated from the final latent states obtained were considered the final plan generated by the model. 

\subsubsection*{Language Inference from Observed Visuo-Proprioceptive Sequences}

The model's ability to understand given visuo-proprioceptive behaviors by inferring corresponding linguistically represented goals was evaluated. Inference of linguistically represented goals was done by minimizing the free energy (See equation 31). In the case of inferring linguistically represented goals, the model generated the linguistic description 5 times by repeated sampling of posterior latent variables ($\mathbf{A}$). The model updates the posterior latent variables ($\mathbf{A}$) and parametric bias ($\mathbf{PB}$) at each iteration of the inference process. Values of $\mathbf{A}$ and linguistic $\mathbf{PB}$ that result in accurate prediction of linguistic goals after 50 iterations of inference were selected as the final result of this inference process for a given visuo-proprioceptive sequence. 

\subsection*{S1: Implementation of Focused Vision Loss}

The vision loss $\mathbf{c}^{\text{att}}_{t}$ performs a scaling of the error signal with respect to the distance from the center of the focused region. Our previous study \cite{jeff2021} showed that this balances the contribution of focused and unfocused regions in the visual prediction. $\mathbf{c}^{\text{att}}_{t}$ is computed as:

\begin{equation}
    \mathbf{c}^{\text{att}}_{t, ij}=\alpha_{\text{s}}\left(
    \begin{bmatrix}\alpha^{\text{att}}_{t,3}\\\alpha^{\text{att}}_{t,4}
    \end{bmatrix}-\begin{bmatrix}\sfrac{i}{\text{N}_{\text{in}}}\\\sfrac{j}{\text{N}_{\text{in}}}
    \end{bmatrix}
    \right)^2,
\end{equation}
for image resolution $(\text{N}_{\text{in}},\text{N}_{\text{in}})$ in order to scale the visual error signal relative to the distance from the center of the focal region.

\subsection*{S2: Variation of Task Compositions in Training}
We use different numbers of combinations to study the model's generalization capability. These are divided into four groups. Group A is a combination of 5 nouns and 8 verbs. Figure S3(A) represents group A1, which used 32/40 (80\%) combinations to train the model. For example the combination of the noun \textit{"Green"} and the verb \textit{"move left"} (reads as \textit{"move green left ."}) is used as a test sequence; therefore, it was not learned by the model during training. 

Figures S3(B) \& (C) show group A2 (60\%) \& A3 (40\%), respectively. Figures S4 (A), (B) and (C) show the compositions used in group B1 (80\%), B2 (60\%) and B3 (40\%); figures S5(A), (B) and (C) show the compositions used in group C1 (80\%), C2 (60\%) and C3 (40\%), and figures S5(D), (E) and (F) show the compositions used in group D1 (7/9, 77\%), D2 (6/9, 66\%) and D3 (3/9, 33\%).

\subsection*{S3: Results}
Here we showcase two sequences generated by the model. Figure S6 shows the visuo-proprioceptive sequence successfully generated by the model to achieve the linguistically specified goal \textit{"put red on green ."}. Figure S7 shows the visuo-proprioceptive sequence generated by the model, where it failed to achieve the linguistically specified goal \textit{"put blue on yellow ."}. 

\subsubsection*{Statistical analysis of model performance}
We performed Welch's unequal variances T-test \cite{welch-t-test} to compare the generalization performance between groups. Welch's T-test was chosen due to the small sample size for training and evaluation. We used a sample size of 5 (number of random seeds used for sampling) and assumed unequal variance for performing the analyses. The results show that group A1 (5x8, 80\%), with the highest number of combinations, performs significantly better than other groups. Table S3 and S4 show a pairwise comparison of the different groups in inferring visuo-proprioceptive trajectories to achieve linguistically represented goals, with training ratios of 80\% and 60\% , respectively.

We also compared the generalization performance for the task of inferring linguistically represented goals from observed visuo-proprioceptive sequences. Even though the overall generalization performance was poor, we observed an increase in performance when the number of compositional elements in training was increased. Table S5 and S6 show a pairwise comparison of the different groups with 80\% and 60\% training ratios, respectively.

\subsection*{S4: Graphical Description of the Inference Process}
Figure S10 illustrates the inference of latent variables for inferring visuo-motor sequences that achieve given linguistically represented goals. Figure S11 illustrates the inference of latent variables for inferring linguistically represented goals from observed visuo-proprioceptive sequences. 

\newpage

\begin{figure}[hbt!]
\centering
\includegraphics[width=0.5\columnwidth]{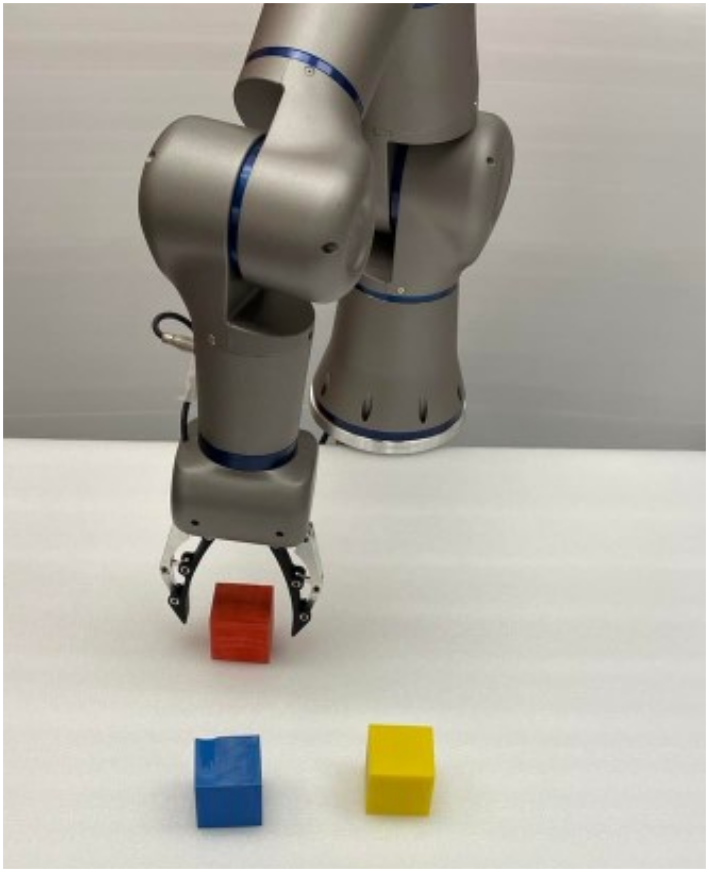}
\caption{Experimental setup: Torobo arm (Tokyo Robotic Inc.) with 7 degrees of freedom manipulating 5 cm cubic blocks of different colors in the workspace.}
\label{exp_setup}
\end{figure}
\newpage

\begin{figure*}[hbt!]
\centering
\includegraphics[width=0.99\textwidth]{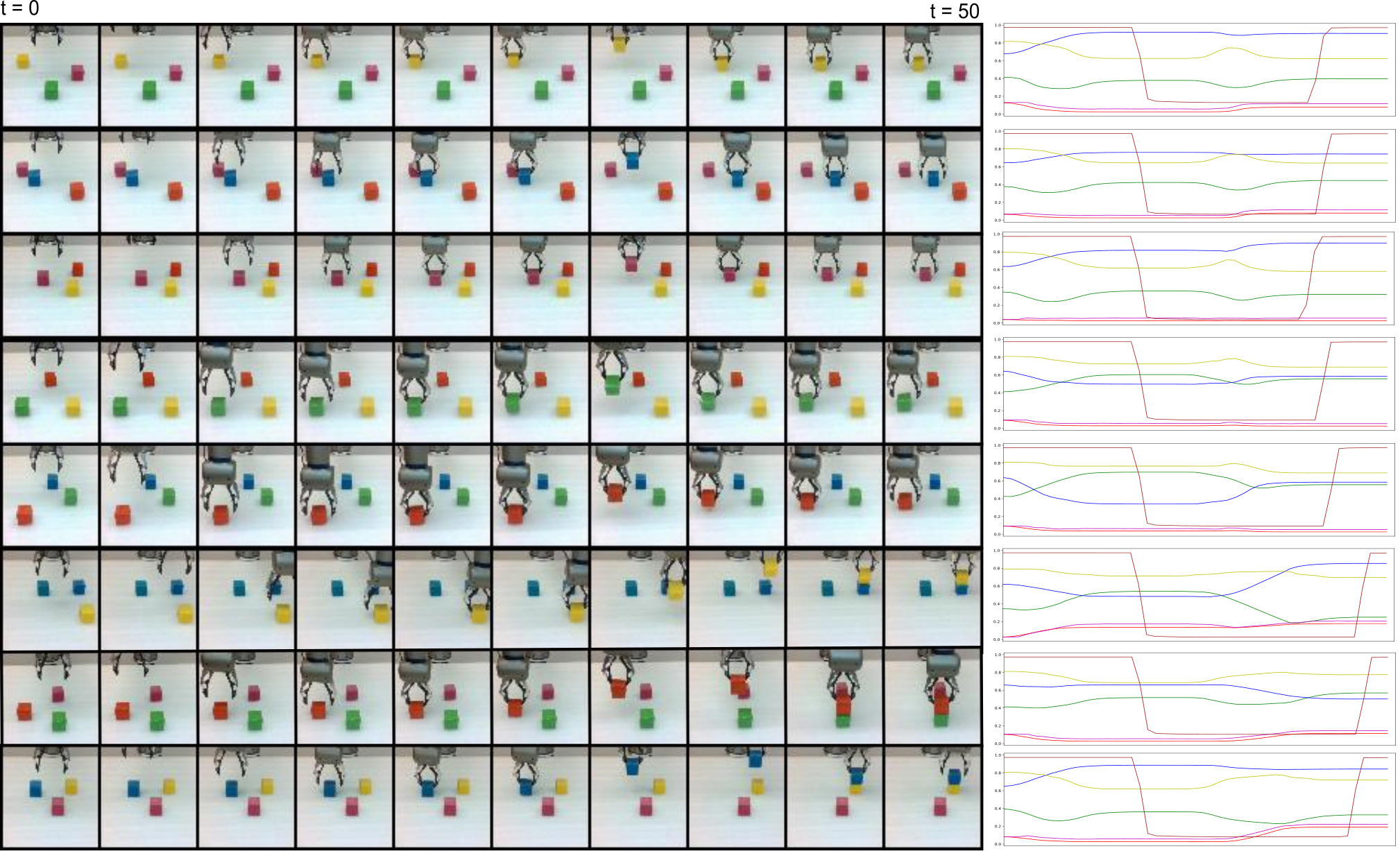}
\caption{Sample visual and corresponding normalized proprioceptive trajectories of each type of task used in the experiment. Top to Bottom: \textit{"grasp yellow"}, \textit{"move blue left"}, \textit{"move purple right"}, \textit{"move green front"}, \textit{"move red back"}, \textit{"put yellow on blue"}, \textit{"put red on green"}, \textit{"put blue on yellow"}. }
\label{training_eg}
\end{figure*}
\newpage

\begin{figure}[hbt!]
\centering
\includegraphics[width=0.99\columnwidth]{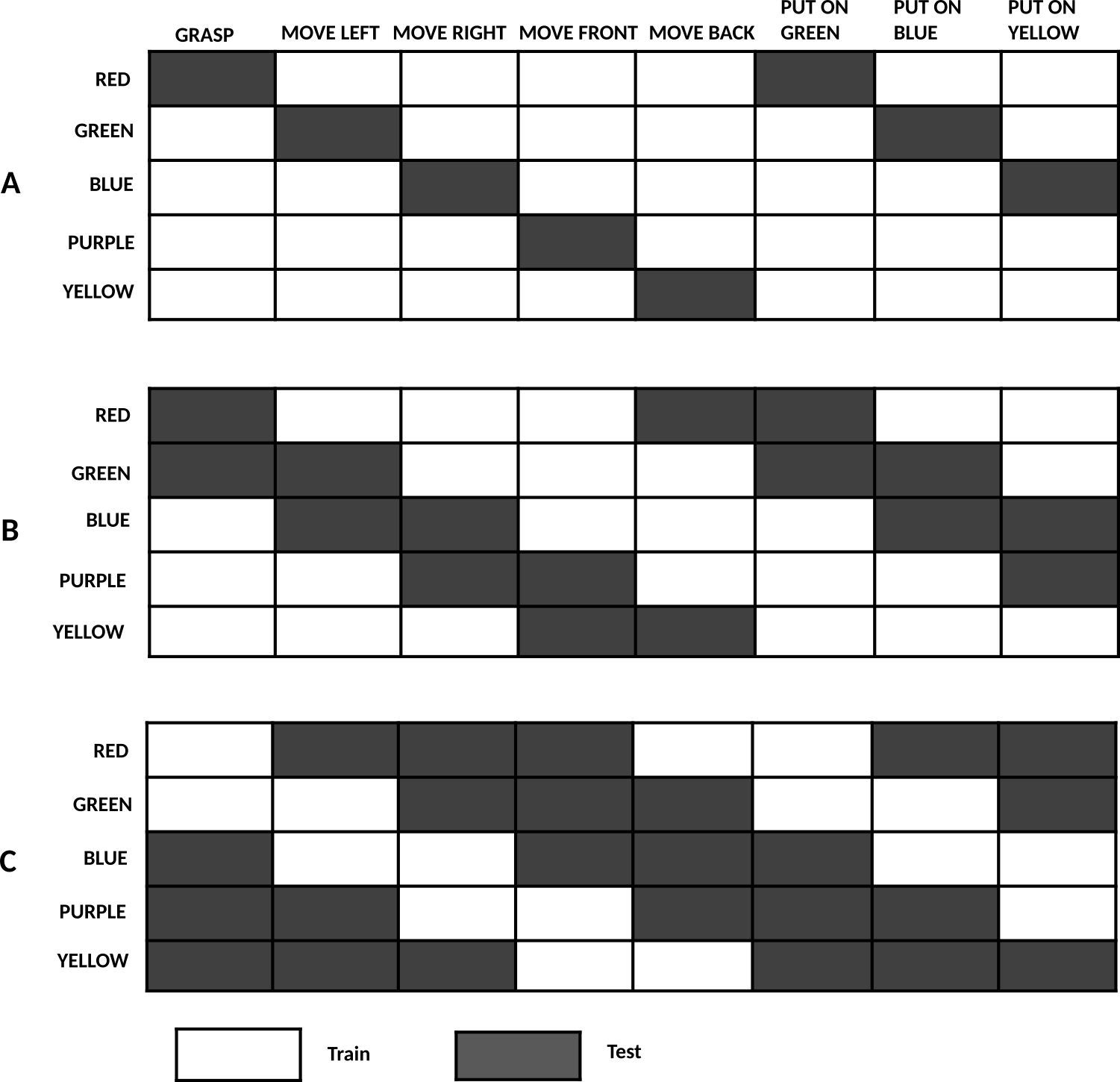}
\caption{Data composition in Group A (5x8). (\textbf{A}) Group A1 (5x8, 80\%), (\textbf{B}) Group A2 (5x8, 60\%), and (\textbf{C}) Group A3 (5x8, 40\%)}
\label{5x8 data}
\end{figure}
\newpage

\begin{figure}[hbt!]
\centering
\includegraphics[width=0.99\columnwidth]{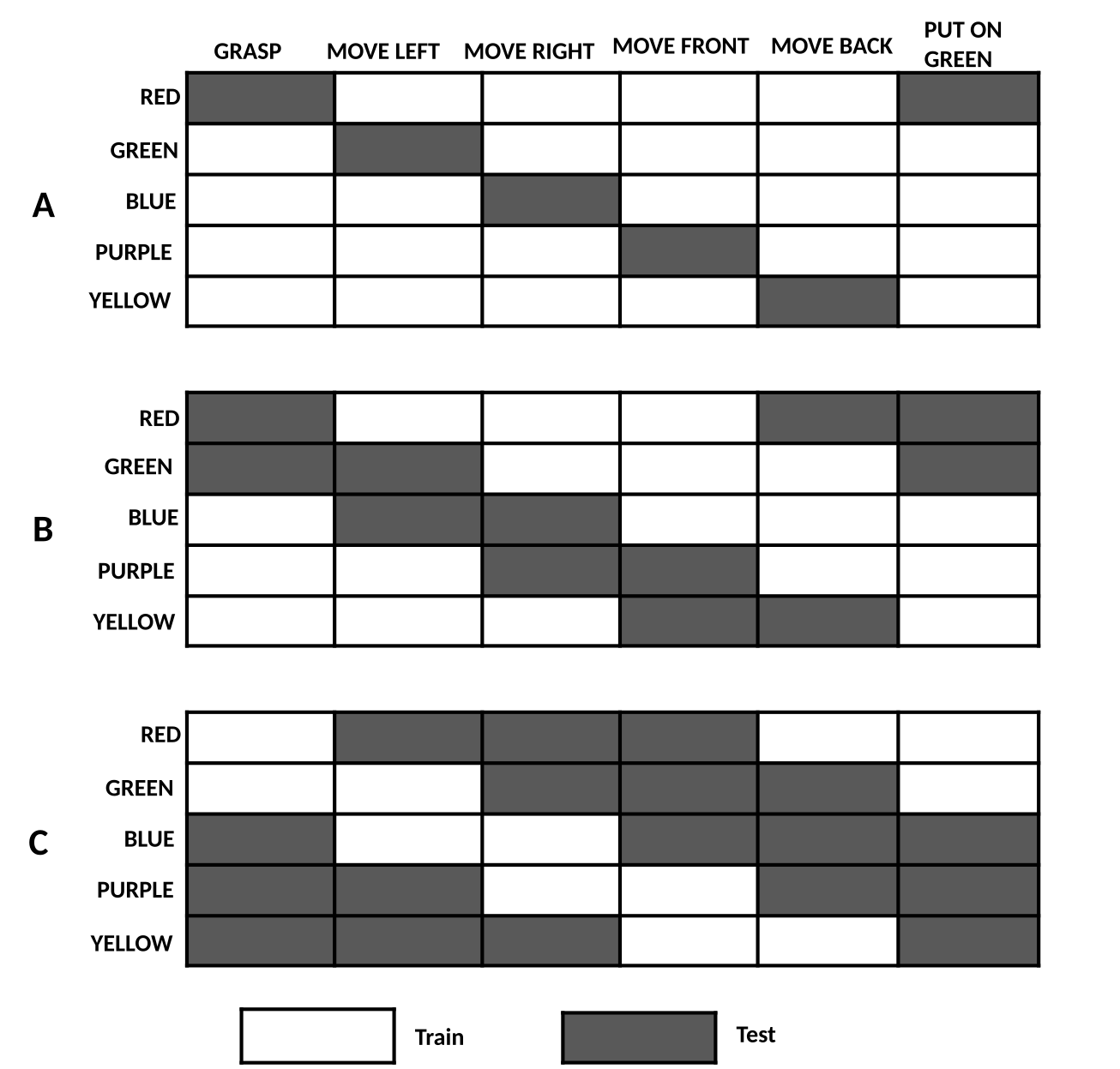}
\caption{Data composition in Group B (5x6). (\textbf{A}) Group B1 (5x6, 80\%), (\textbf{B}) Group B2 (5x6, 60\%), and (\textbf{C}) Group B3 (5x6, 40\%)}
\label{5x6 data}
\end{figure}
\newpage

\begin{figure}[hbt!]
\centering
\includegraphics[width=0.99\columnwidth]{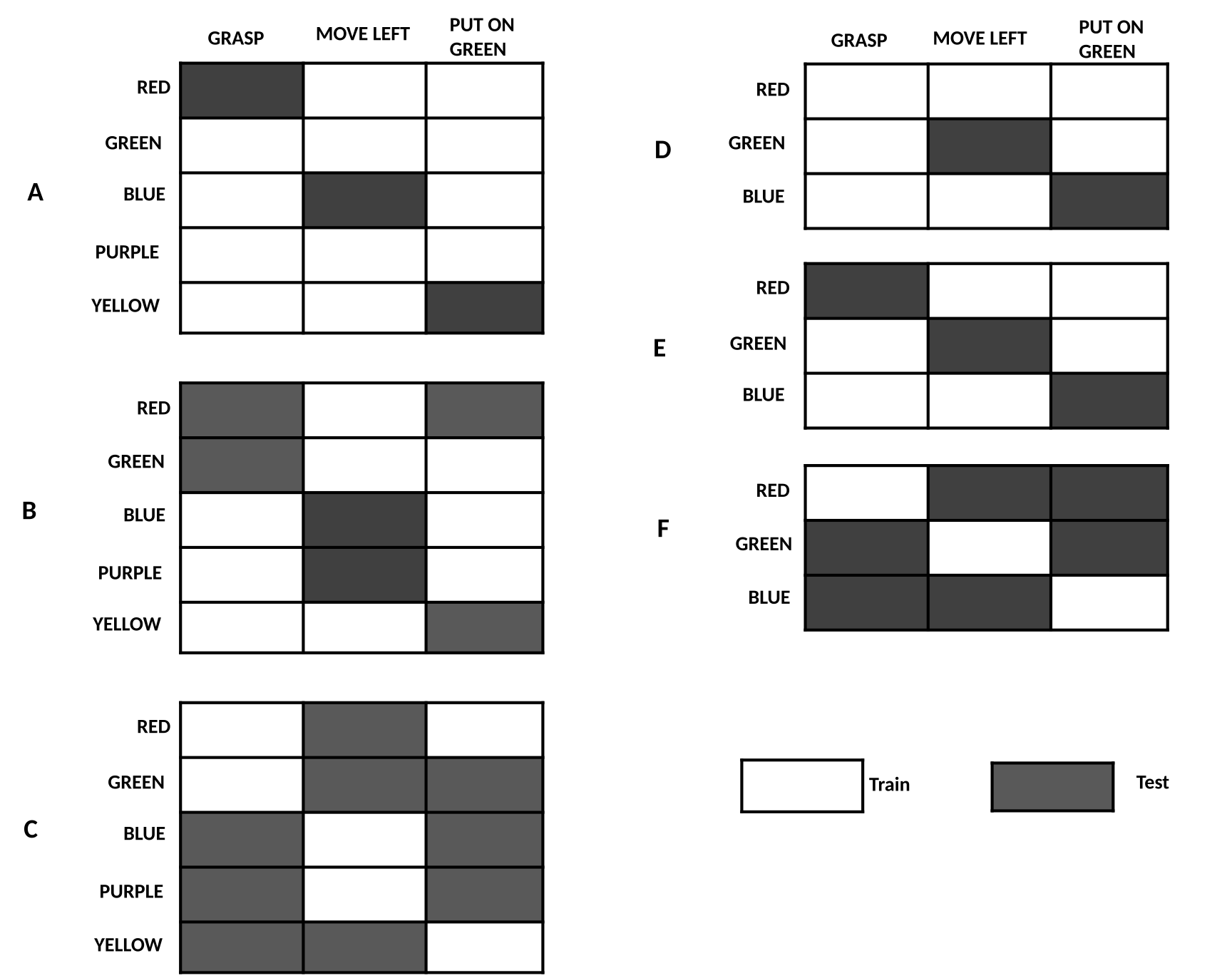}
\caption{Data composition in Group C and D (5x3 and 3x3). (\textbf{A}) Group C1 (5x3, 80\%), (\textbf{B}) Group C2 (5x3, 60\%), (\textbf{C}) Group C3 (5x3, 40\%), (\textbf{D}) Group D1 (3x3, 77\%), (\textbf{E}) Group D2 (3x3, 66\%), and (\textbf{F}) Group D3 (3x3, 33\%)}
\label{5x3 3x3 data}
\end{figure}
\newpage

\begin{figure*}[hbt!]
\centering
\includegraphics[width=0.99\textwidth]{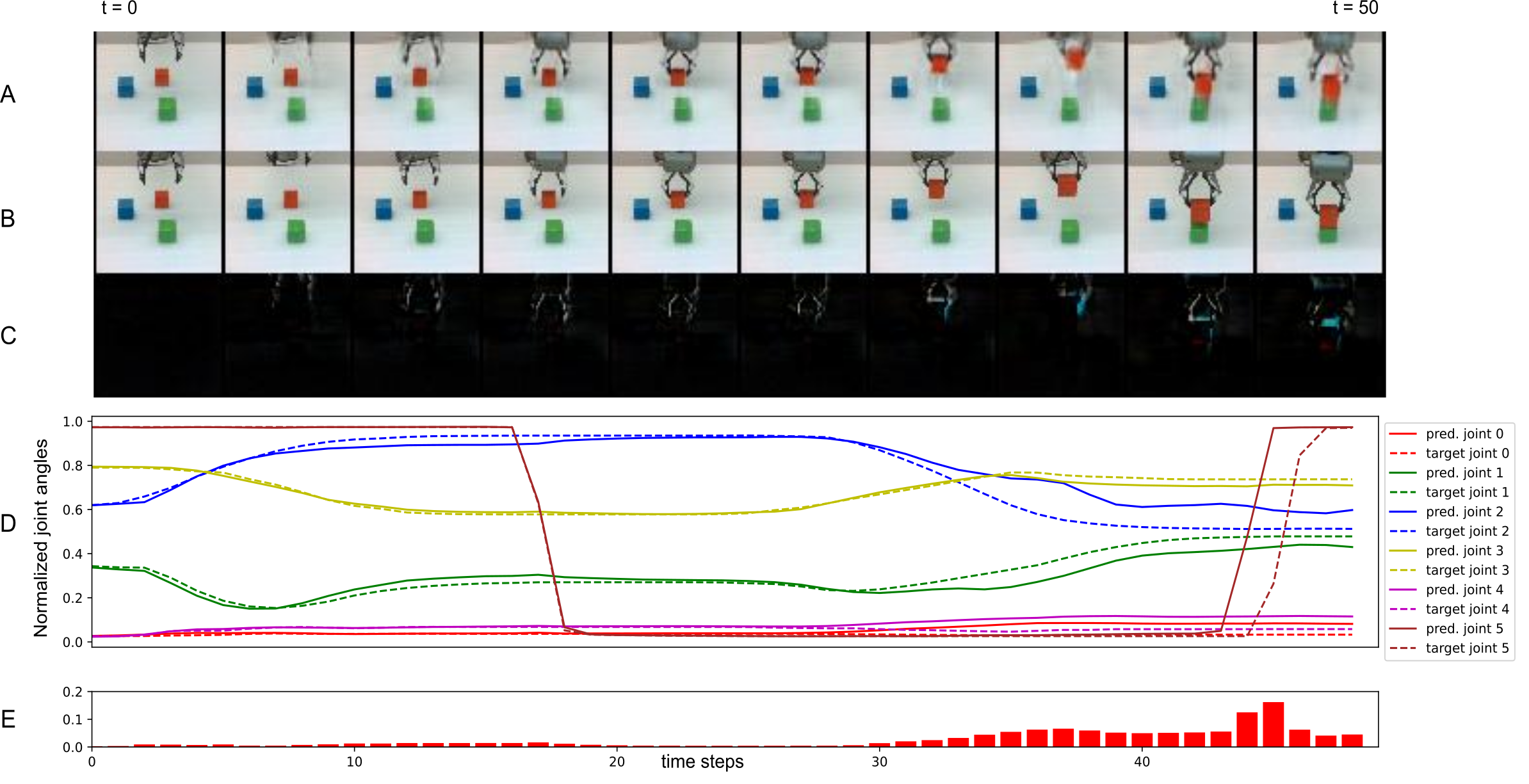}
\caption{Visuo-Proprioceptive sequence generated by the model to achieve the linguistically represented goal of \textit{"put red on green."}. (\textbf{A}): imagined visual trajectory generated by the model; (\textbf{B}): the ground truth target; (\textbf{C}): difference between predicted and visual sequence and ground truth, (\textbf{D}): normalized joint angle trajectory predicted by the model compared with the corresponding ground truth; (\textbf{E}): the mean difference between the predicted joint angles and the ground truth.}
\label{svis_example}
\end{figure*}
\newpage

\begin{figure*}[hbt!]
\centering
\includegraphics[width=0.99\textwidth]{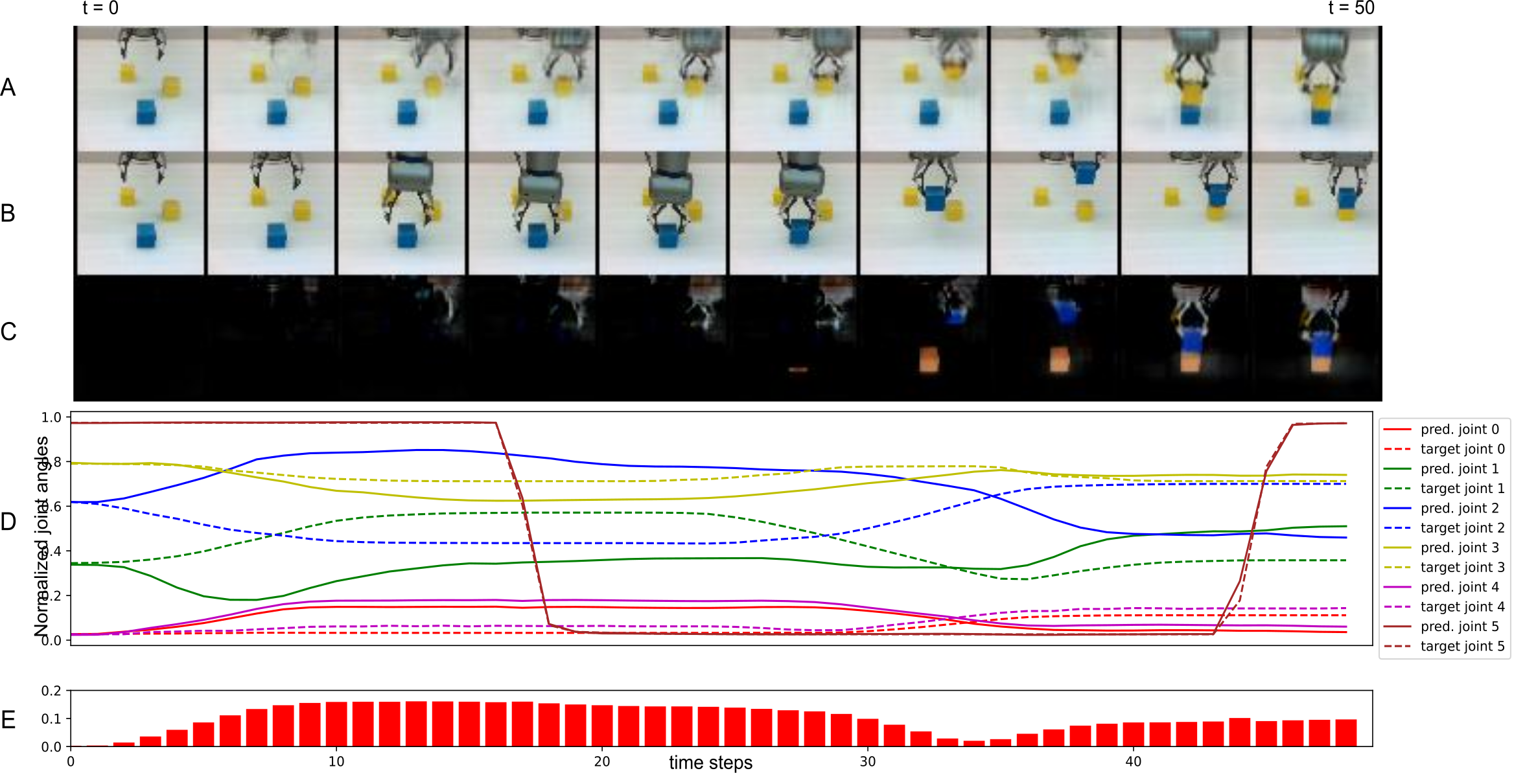}
\caption{Visuo-Proprioceptive sequence generated by the model to achieve the linguistically represented goal of \textit{"put blue on yellow."}. (\textbf{A}): imagined visual trajectory generated by the model; (\textbf{B}): the ground truth target; (\textbf{C}): difference between predicted and visual sequence and ground truth, (\textbf{D}): normalized joint angle trajectory predicted by the model compared with the corresponding ground truth; (\textbf{E}): the mean difference between the predicted joint angles and the ground truth.}
\label{failvis_example}
\end{figure*}
\newpage

\begin{figure*}[hbt!]
\centering
\includegraphics[width=0.99\textwidth]{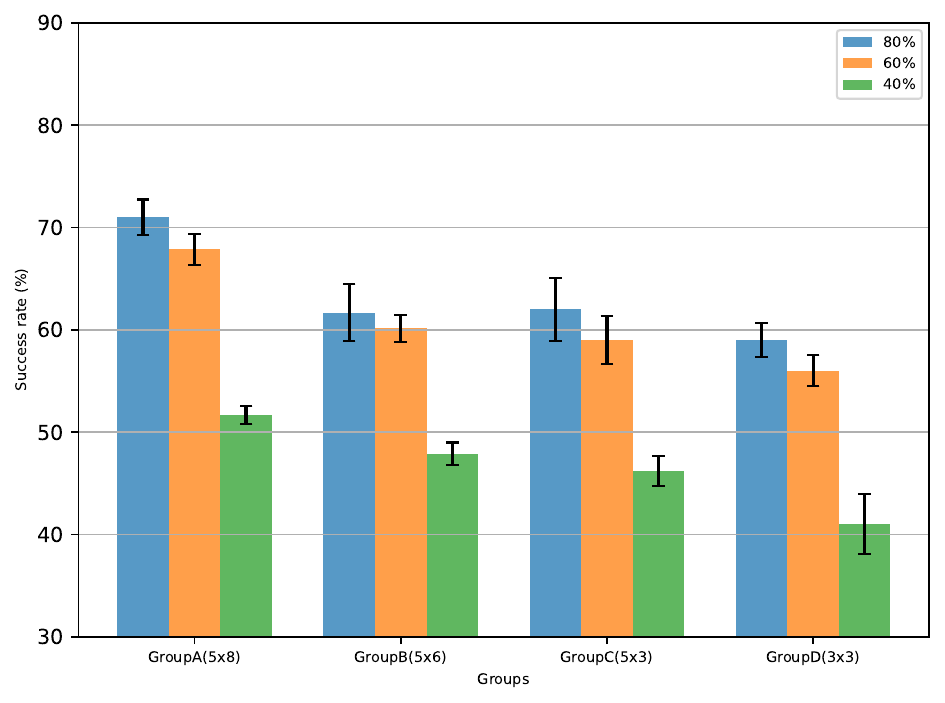}
\caption{Comparison of generalization performance, for inference of linguistic goals, for unlearned compositions (U-C) among groups with different number of compositions and training ratios.}
\label{group1_success rate}
\end{figure*}
\newpage

\begin{figure*}[hbt!]
\centering
\includegraphics[width=0.99\textwidth]{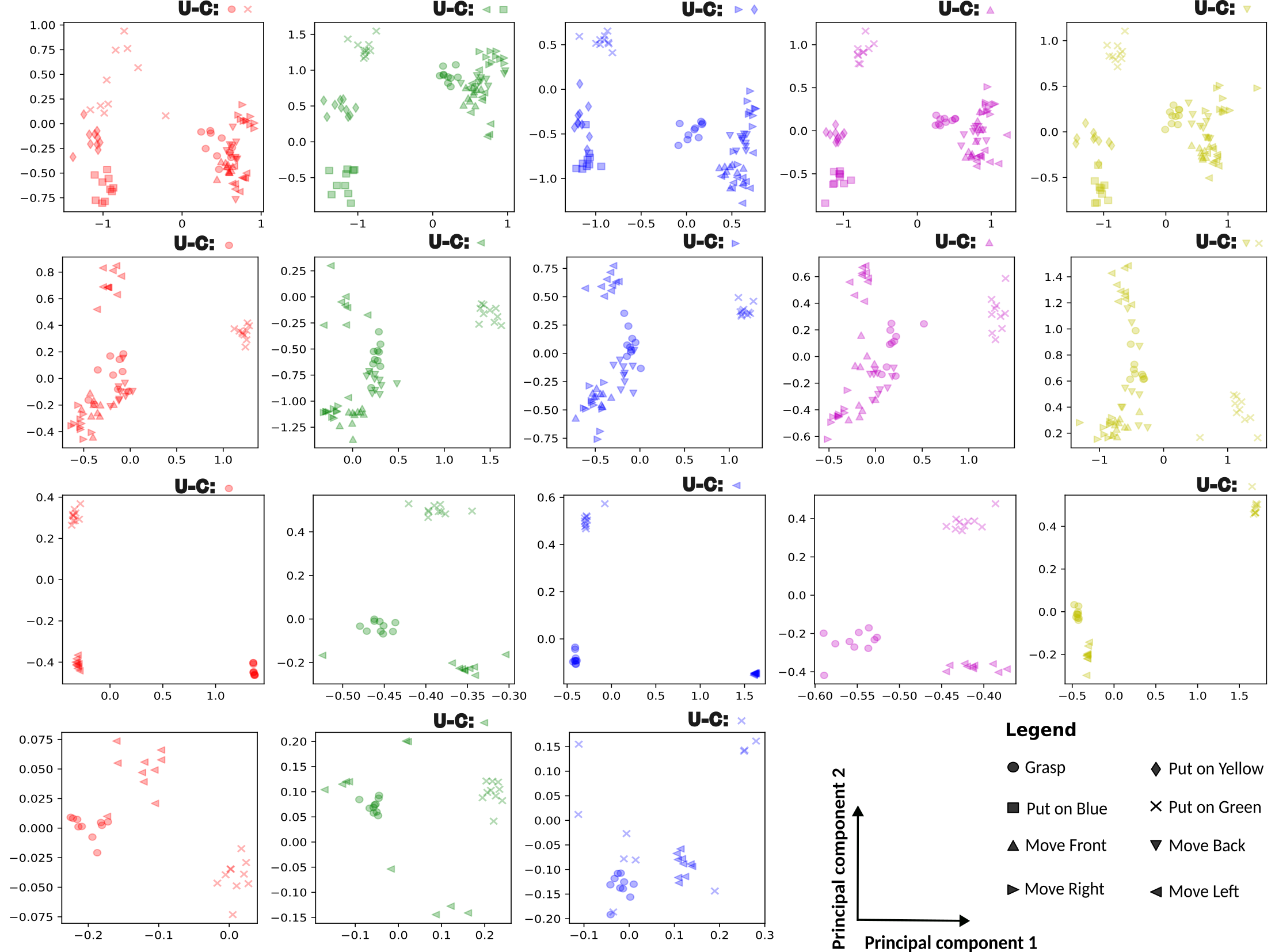}
\caption{Scatterplot of Kernel PCA of the latent state PB vector for all groups with highest training ratio. (\textbf{A}): Group A1 (5x8, 80\%) (\textbf{B}): Group B1 (5x6, 80\%) (\textbf{C}): Group C1 (5x3, 80\%) (\textbf{D}): Group D1 (3x3, 77\%). U-C refers to unlearned compositions that were used for testing. Colors of markers indicate colors of objects being manipulated. The variance explained by the two components of KPCA for all groups was greater than 90\%}
\label{kpca_pb_80}
\end{figure*}
\newpage

\begin{figure}[hbt!]
\centering
\includegraphics[width=1\textwidth]{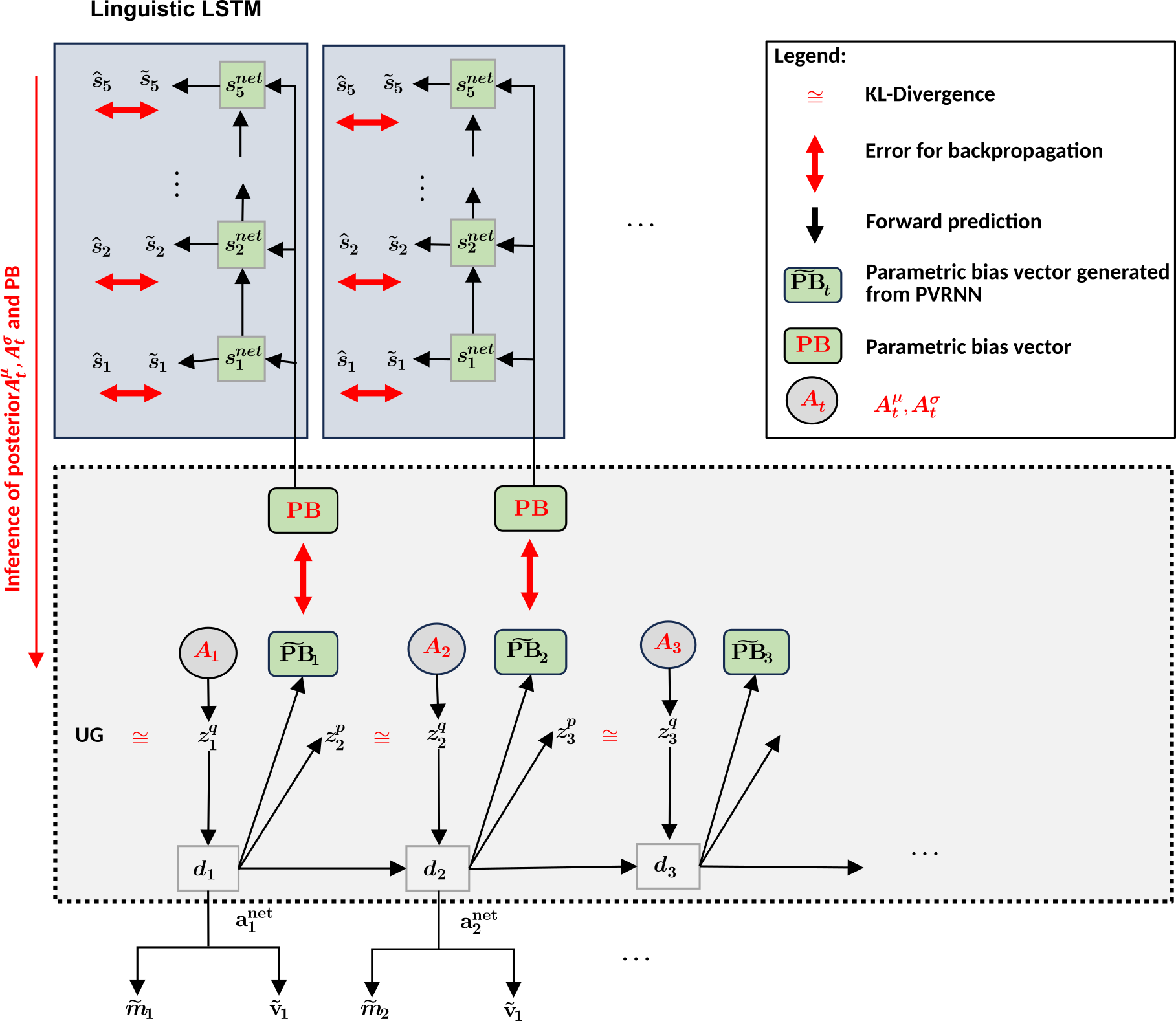}
\caption{Graphical description of goal-directed planning using active inference to infer visuo-proprioceptive sequences given linguistically represented goal. Parameters in red text are updated in order to minimize the prediction error}
\label{AIF}
\end{figure}
\newpage

\begin{figure}[hbt!]
\centering
\includegraphics[width=1\textwidth]{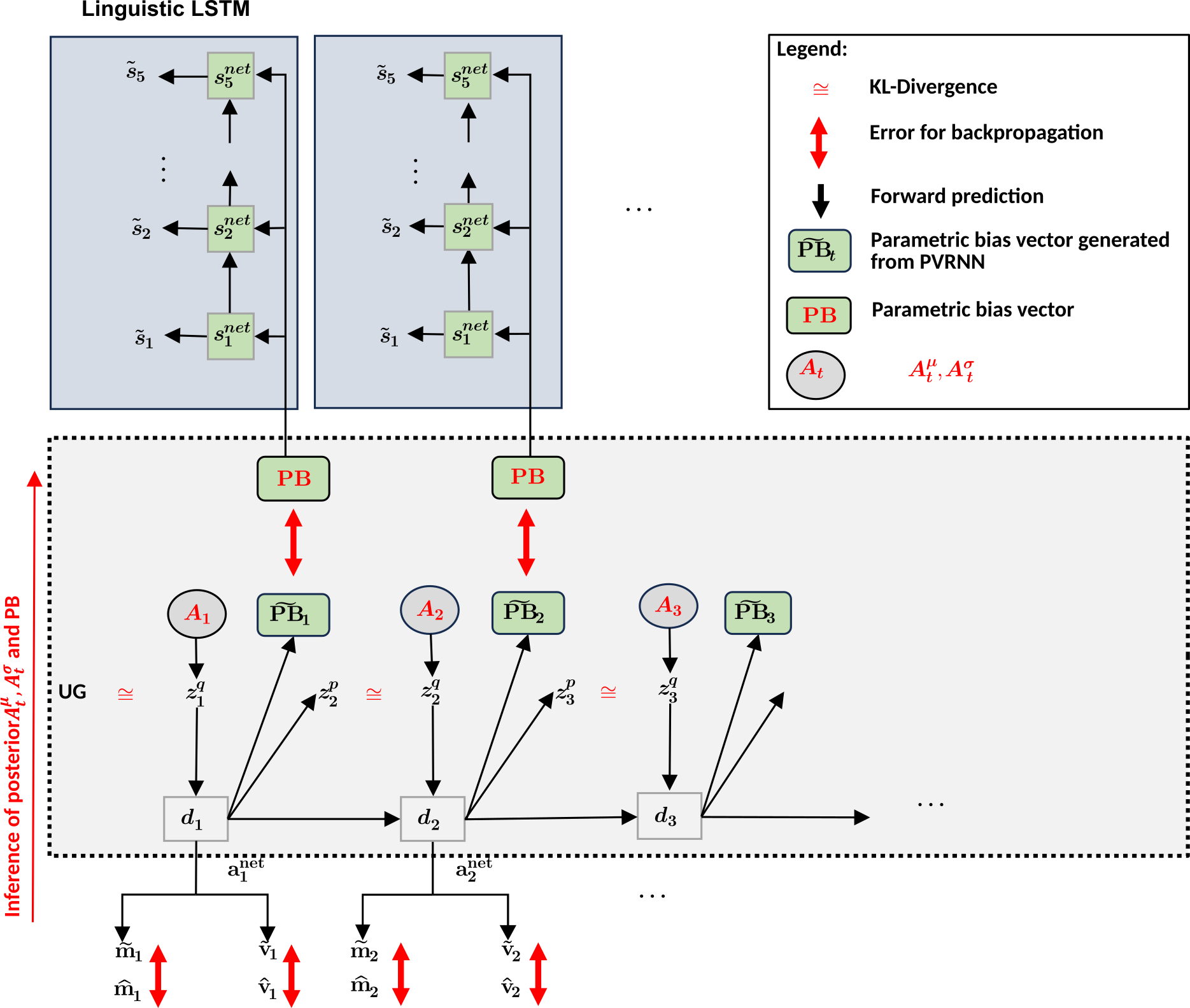}
\caption{Graphical description of inferring linguistically represented goal from observed visuo-proprioceptive sequences. Parameters in red text are updated in order to minimize the prediction error}
\label{goal_inf}
\end{figure}
\newpage

\begin{table}
    \centering
    \caption{Accuracy: Inference of Visuo-Proprioceptive Sequences}
    \begin{tabular}{c cc}
        \hline
            & \multicolumn{2}{c}{\bf{Visuo-proprioceptive Error ($\mu \pm{SD}$) \%}} \\
        \hline
            &\bf{U-P}  &\bf{U-C} \\
        \hline
            \bf{Group A1 (5x8, 80\%)}   &$0.0306\pm{0.0016}$    &$0.0361\pm{0.0049}$    \\ 
            \bf{Group A2 (5x8, 60\%)}   &$0.0331\pm{0.0017}$    &$0.0500\pm{0.0021}$    \\
            \bf{Group A3 (5x8, 40\%)}   &$0.0342\pm{0.0018}$    &$0.0710\pm{0.0080}$    \\
            \bf{Group B1 (5x6, 80\%)}   &$0.0328\pm{0.0015}$    &$0.0494\pm{0.0040}$    \\
            \bf{Group B2 (5x6, 60\%)}   &$0.0334\pm{0.0017}$    &$0.0597\pm{0.0065}$    \\
            \bf{Group B3 (5x6, 40\%)}   &$0.0346\pm{0.0019}$    &$0.0890\pm{0.0054}$    \\
            \bf{Group C1 (5x3, 80\%)}   &$0.0346\pm{0.0018}$    &$0.0615\pm{0.0031}$    \\
            \bf{Group C2 (5x3, 60\%)}   &$0.0351\pm{0.0021}$    &$0.0791\pm{0.0059}$    \\
            \bf{Group C3 (5x3, 40\%)}   &$0.0395\pm{0.0010}$    &$0.1071\pm{0.0057}$    \\
            \bf{Group D1 (3x3, 77\%)}   &$0.0341\pm{0.0018}$    &$0.0655\pm{0.0053}$    \\
            \bf{Group D2 (3x3, 66\%)}   &$0.0344\pm{0.0019}$    &$0.0896\pm{0.0054}$    \\
            \bf{Group D3 (3x3, 33\%)}   &$0.0405\pm{0.0021}$    &$0.1226\pm{0.0093}$    \\
        \hline
    \end{tabular}
    \label{table1}
\end{table}

\begin{table}
    \centering
    \caption{Accuracy: Inference of Linguistically Represented Goals}
    \begin{tabular}{c cc}
        \hline
            & \multicolumn{2}{c}{\bf{Success rate ($\mu \pm{SD}) \%$}} \\
        \hline
            &\bf{U-P}  &\bf{U-C}\\
        \hline
            \bf{Group A1 (5x8, 80\%)}   &$72.50\pm{0.71}$    &$71.00\pm{1.74}$\\
            \bf{Group A2 (5x8, 60\%)}   &$69.16\pm{0.69}$    &$67.87\pm{1.51}$\\
            \bf{Group A3 (5x8, 40\%)}   &$70.00\pm{1.42}$    &$51.66\pm{0.88}$\\
            \bf{Group B1 (5x6, 80\%)}   &$70.00\pm{0.74}$    &$61.66\pm{2.78}$\\
            \bf{Group B2 (5x6, 60\%)}   &$70.55\pm{0.99}$    &$60.16\pm{1.32}$\\
            \bf{Group B3 (5x6, 40\%)}   &$68.33\pm{0.91}$    &$47.88\pm{1.11}$\\
            \bf{Group C1 (5x3, 80\%)}   &$71.66\pm{1.39}$    &$62.00\pm{3.06}$\\
            \bf{Group C2 (5x3, 60\%)}   &$71.11\pm{0.99}$    &$59.00\pm{2.38}$\\
            \bf{Group C3 (5x3, 40\%)}   &$68.33\pm{2.78}$    &$46.22\pm{1.49}$\\
            \bf{Group D1 (3x3, 77\%)}   &$71.42\pm{2.02}$    &$59.00\pm{1.67}$\\
            \bf{Group D2 (3x3, 66\%)}   &$71.66\pm{1.82}$    &$56.00\pm{1.52}$\\
            \bf{Group D3 (3x3, 33\%)}   &$70.00\pm{2.98}$    &$41.00\pm{2.92}$\\
        \hline
    \end{tabular}
    \label{table3}
\end{table}

\begin{table}
    \centering
    \caption{Welch's T-test: Inference of Visuo-Proprioceptive Sequences,\\ Training Ratio: 80\%}
    \begin{tabular}{cccc}
        \hline
            \bf{Group 1}   &\bf{Group 2}  &\bf{T-statistic} &\bf{P-value} \\
        \hline
            \bf{A1}   &\bf{B1}  &-7.43    &9.03x$10^{-5}$     \\ 
            \bf{A1}   &\bf{C1}  &-15.48    &1.56x$10^{-6}$     \\ 
            \bf{A1}   &\bf{D1}  &-14.48    &5.60x$10^{-7}$     \\ 
            \bf{B1}   &\bf{C1}  &-8.45    &4.19x$10^{-5}$   \\ 
            \bf{B1}   &\bf{D1}  &-8.57    &4.09x$10^{-5}$    \\ 
            \bf{C1}   &\bf{D1}  &-2.30    &5.78x$10^{-2}$    \\ 
        \hline
    \end{tabular}
    \label{vp80welcht-test}
\end{table}

\begin{table}
    \centering
    \caption{Welch's T-test: Inference of Visuo-Proprioceptive Sequences,\\ Training Ratio: 60\%}
    \begin{tabular}{cccc}
        \hline
            \bf{Group 1}   &\bf{Group 2}  &\bf{T-statistic} &\bf{P-value} \\
        \hline
            \bf{A2}   &\bf{B2}  &-5.02  &4.44x$10^{-3}$    \\ 
            \bf{A2}   &\bf{C2}  &-16.42  &1.50x$10^{-5}$     \\ 
            \bf{A2}   &\bf{D2}  &-24.16  &2.00x$10^{-6}$     \\ 
            \bf{B2}   &\bf{C2}  &-7.81  &5.40x$10^{-5}$     \\ 
            \bf{B2}   &\bf{D2}  &-12.50  &2.00x$10^{-6}$   \\ 
            \bf{C2}   &\bf{D2}  &-4.64  &1.69x$10^{-3}$     \\ 
        \hline
    \end{tabular}
    \label{vp60welcht-test}
\end{table}

\begin{table}
    \centering
    \caption{Welch's T-test: Inference of Linguistically Represented Goal,\\ Training Ratio: 80\%}
    \begin{tabular}{ccccc}
        \hline
            \bf{Group 1}   &\bf{Group 2}  &\bf{T-statistic} &\bf{P-value}  \\
        \hline
            \bf{A1}   &\bf{B1}  &10.06  &2.6x$10^{-5}$      \\ 
            \bf{A1}   &\bf{C1}  &9.03  &7.4x$10^{-5}$     \\ 
            \bf{A1}   &\bf{D1}  &17.59  &1.1x$10^{-7}$     \\ 
            \bf{B1}   &\bf{C1}  &-0.29  &7.7x$10^{-1}$   \\ 
            \bf{B1}   &\bf{D1}  &2.89  &2.4x$10^{-2}$    \\ 
            \bf{C1}   &\bf{D1}  &3.04 &2.1x$10^{-2}$     \\ 
        \hline
    \end{tabular}
    \label{langinf80welcht-test}
\end{table}

\begin{table}
    \centering
    \caption{Welch's T-test: Inference of Linguistically Represented Goal,\\ Training Ratio: 60\%}
    \begin{tabular}{cccc}
        \hline
            \bf{Group 1}   &\bf{Group 2}  &\bf{T-statistic} &\bf{P-value} \\
        \hline
            \bf{A2}   &\bf{B2}  &13.59  &9.73 x $10^{-7}$     \\ 
            \bf{A2}   &\bf{C2}  &11.12  &1.34 x $10^{-5}$     \\ 
            \bf{A2}   &\bf{D2}  &17.59  &1.13 x $10^{-7}$     \\ 
            \bf{B2}   &\bf{C2}  &1.50  &1.80 x $10^{-1}$     \\ 
            \bf{B2}   &\bf{D2}  &7.30  &9.20 x $10^{-5}$     \\ 
            \bf{C2}   &\bf{D2}  &3.75  &7.50 x $10^{-3}$      \\ 
        \hline
    \end{tabular}
    \label{langinf60welcht-test}
\end{table}
\newpage

\begin{algorithm}[H]
\caption{Training}\label{alg:alg1}
Initialize parameters:\\
$\mathbf{\theta_{net}}, \mathbf{A^d}, \mathbf{PB^d} ; \forall_d \in \mathcal{D}_{train}$\\ 
  \For{$e\leftarrow 1$ \KwTo $N_{\text{epochs}}$}{
   \For{$t\leftarrow 1$ \KwTo $\mathbf{T}$}{
        sample posterior:\\
        $\mathbf{z}^q_{t} \leftarrow \B{\mu}_{t}^q +\B{\epsilon}*\B{\sigma}_{t}^q$\!\\
        generation:\\
     $(\mathbf{\widetilde{v}}_{t},\mathbf{\widetilde{m}}_{t}, \mathbf{\widetilde{S}},\mathbf{\widetilde{PB}}_{t})\longleftarrow\\ Fwd(\mathbf{\widetilde{v}_{t-1}},\mathbf{\widetilde{m}_{t-1}},\mathbf{\widetilde{S},\mathbf{PB^d},\mathbf{z}^q_{t}},\mathbf{\theta_{net}})$\\
     }     
    compute loss:\\
   $l \leftarrow L(\mathbf{\widetilde{v}},\mathbf{\widetilde{m}}, \mathbf{\widetilde{S}},\mathbf{\widetilde{PB}^d}, \mathbf{{v}},\mathbf{{m}}, \mathbf{{S}},\mathbf{PB^d},)$\;\\
   gradient descent:\\
   $({\mathbf{\theta_{net}}},{\mathbf{A^d}, \mathbf{PB}}) \leftarrow Adam(\partial_{\B{\theta}}l,\partial_{\B{A}}l, \partial_{\B{PB}}l)$;\\
     }
\end{algorithm}

\begin{algorithm}[H]
\caption{Inference of visuo-motor sequence}\label{alg:alg2}
Initialize parameters:\\
$\mathbf{PB}^g,  \mathbf{{A^g}} ;  \forall_g \in \mathcal{D}_{test}$\\ 
  \For{$e\leftarrow 1$ \KwTo $N_{\text{iterations}}$}{
   \For{$t\leftarrow 1$ \KwTo $\mathbf{T}$}{
        sample posterior:\\
        $\mathbf{z}^{q}_{t}\leftarrow \B{\mu}_{t}^{q} +\B{\epsilon}\times\B{\sigma}_{t}^{q}$\;\\
        generation:\\
     $(\mathbf{\widetilde{v}}_{t},\mathbf{\widetilde{m}}_{t},\mathbf{\widetilde{PB}}_{t}, \mathbf{\widetilde{S}})\longleftarrow\\ Fwd(\mathbf{\widetilde{v}_{t-1}},\mathbf{\widetilde{m}_{t-1}},\mathbf{\widetilde{S}},\mathbf{PB^g},\mathbf{z_{t}^{q}},\mathbf{\theta_{net}})$\\
     }
    loss calculation:\\
   $l^g \leftarrow L^g(\mathbf{\widetilde{S}^g},\mathbf{\widetilde{PB}}^g, \mathbf{{S}^g},\mathbf{{PB}}^g)$\;\\
   gradient descent:\\
   $({\mathbf{A}^g, \mathbf{PB}^g})\leftarrow Adam(\partial_{\B{A}}l^g, \partial_{\B{PB}}l^g)$;\\
     }
    \text{Choose best solution from all iterations}
\end{algorithm}

\begin{algorithm}[H]
\caption{Inference of linguistic goals}\label{alg:alg3}
Initialize parameters:\\
$\mathbf{PB}^g,  \mathbf{{A^g}} ;  \forall_g \in \mathcal{D}_{test}$\\ 
  \For{$e\leftarrow 1$ \KwTo $N_{\text{iterations}}$}{
   \For{$t\leftarrow 1$ \KwTo $\mathbf{T}$}{
        sample posterior:\\
        $\mathbf{z}^{q}_{t}\leftarrow \B{\mu}_{t}^{q} +\B{\epsilon}\times\B{\sigma}_{t}^{q}$\;\\
        generation:\\
     $(\mathbf{\widetilde{v}}_{t},\mathbf{\widetilde{m}}_{t},\mathbf{\widetilde{PB}}_{t}, \mathbf{\widetilde{S}})\longleftarrow\\ Fwd(\mathbf{\widetilde{v}_{t-1}},\mathbf{\widetilde{m}_{t-1}},\mathbf{\widetilde{S}},\mathbf{PB^g},\mathbf{z_{t}^{q}},\mathbf{\theta_{net}})$\\
     }
    loss calculation:\\
   $l^g \leftarrow L^g (\mathbf{\widetilde{v}},\mathbf{\widetilde{m}},\mathbf{\widetilde{PB}^g}, \mathbf{{v}},\mathbf{{m}}, \mathbf{PB^g})$\;\\
   gradient descent:\\
   $({\mathbf{A}^g, \mathbf{PB}^g})\leftarrow Adam(\partial_{\B{A}}l^g, \partial_{\B{PB}}l^g)$;\\
     }
    \text{Choose best solution from all iterations}
\end{algorithm}

\end{document}